\documentclass{article}

\usepackage[T1]{fontenc}
\usepackage{times}
\usepackage{hyperref}
\usepackage{url}
\usepackage{color}  
\hypersetup{urlcolor=black,linkcolor=black,citecolor=black,colorlinks=true} 
\usepackage{amsmath}
\usepackage{mathabx}
\usepackage{dcolumn}
\usepackage{graphics}
\DeclareGraphicsExtensions{.pdf,.png,.jpg}
\usepackage{graphicx}
\usepackage{verbatim}
\usepackage{afterpage}
\usepackage{multirow}
\usepackage{subcaption}
\usepackage{stfloats}
\usepackage{todonotes}
\usepackage{wasysym}

\newcommand{\surl}[1]{{\small\url{#1}}}

\def\-{\discretionary{}{}{}} 

\newcolumntype{d}[1]{D{.}{.}{#1}}

\begin{document}
\title{A visual approach for age and gender identification on Twitter\thanks{Preprint of~\cite{thispaper}. The final publication is available at IOS Press through \url{http://dx.doi.org/10.3233/JIFS-169497}}}

\author{Miguel A. {A}lvarez-Carmona$^1$, Luis Pellegrin$^1$, Manuel Montes-y-G\'{o}mez$^1$,\\ Fernando S\'{a}nchez-Vega$^1$, Hugo Jair Escalante$^1$, A. Pastor L\'{o}pez-Monroy$^2$, \\ Luis Villase\~{n}or-Pineda$^1$, Esa\'{u} Villatoro-Tello$^3$ \\
$^1$\emph{Computer Science Department}\\ \emph{Instituto Nacional de Astrof\'{i}sica, \'{O}ptica y Electr\'{o}nica} \\ \emph{Luis Enrique Erro 1, Puebla 72840, M\'{e}xico}\\
$^2$ \emph{Research in Text Understanding and Analysis of Language Lab}\\ \emph{University of Houston} \\4800 Calhoun Road, Houston, TX 77004, USA\\
$^3$ \emph{Language and Reasoning Research Group}\\\emph{Information Technologies Department, Universidad Aut\'{o}noma Metropolitana}\\ \emph{Unidad Cuajimalpa (UAM-C), Ciudad de M\'{e}xico, 05348, M\'{e}xico}}
\maketitle


\vspace{-7mm}
\begin{abstract}
The goal of Author Profiling (AP) is to identify  demographic aspects (e.g., age, gender) from a given set of authors by analyzing their written texts. Recently, the AP task has gained interest in many problems related to computer forensics, psychology, marketing, but specially in those related with social media exploitation. As known, social media data is shared through a wide range of modalities (e.g., text, images and audio), representing valuable information to be exploited for extracting valuable insights from users. Nevertheless, most of the current work in AP using social media data has been devoted to analyze textual information only, and there are very few works that have started exploring the gender identification using visual information. Contrastingly, this paper focuses in exploiting the visual modality to perform both age and gender identification in social media, specifically in Twitter. Our goal is to evaluate the pertinence of using visual information in solving the AP task. Accordingly, we have extended the Twitter corpus from PAN 2014, incorporating posted images from all the users, making a distinction between tweeted and retweeted images. Performed experiments provide interesting evidence on the usefulness of visual information in comparison with traditional textual representations for the AP task. 
\end{abstract}


\section{Introduction}
Nowadays there is a tremendous amount of information available on the Internet. Specifically, social media domains are constantly growing thanks to the information generated by a huge community of active users. Such information is available in several modalities, including  text, image, audio and video. The availability of all this information plays an important role in designing appropriate tools for diverse tasks and applications. Particularly, during recent years, the Author Profiling (AP) task has gained interest among the scientific community. AP aims at revealing demographic information (e.g., age, gender, native language, personality traits, cultural background) of authors through analyzing their written texts~\cite{koppel2002automatically}. The AP task has a wide range of applications and it could have a broad impact in a number of problems. For instance, in forensics, profiling authors could be used as valuable additional evidence; in marketing, this information could be exploited to improve targeted advertising. 

As known, social media data has a multimodal nature (e.g., text, images, audio, social interactions), however, most of the previous research on AP has been devoted to the analysis of the textual modality \cite{burger2011discriminating,koppel2002automatically,nguyen2013old,peersman2011predicting,schler2006effects}, disregarding information from the other modalities that could be potentially useful for improving the performance of AP methods. Accordingly, some works have begun to exploit distinct modalities for approaching the AP problem~\cite{Can13,Merler15,taniguchi,You13}. The visual modality has resulted particularly interesting, mostly because it is, to some extent, language independent nature. In fact, previous work has found a relationship between images and users' interests, opinions and thoughts~\cite{Cristani13,Eftekhar14,Hum11,Wu15,YangHE15,You16,You14}.

Although visual information is particularly appealing for AP, it is just recently that some authors began to pay attention to the content of images shared by users. For example, for gender identification, some authors have exploited the information provided by the colors adopted by users in their profiles~\cite{Lovato14}. In~\cite{Azam16} authors used state of the art face-gender recognizers over user profile pictures. Nonetheless, the most common strategy so far consist in exploiting the \textit{posting behavior}, which implies the manual classification of posted images in order to analyze histograms of classes/objects posted by users~\cite{Hum11}. 

Despite previous efforts for including visual information in the AP task, only the gender recognition problem has been studied, leaving the age identification problem unexplored. In addition, many of the previous research considers an scenario where manually tagged images are provided for training, resulting in an impractical and unrealistic scenario for AP systems. 

In order to overcome these limitations, we present a thorough analysis on the pertinence of visual information for approaching the AP problem, targeting both, age and gender identification. Our study comprises an analysis on the discriminative capabilities of tweeted and retweeted images by users. As part of the study, a method for AP using images is proposed in this paper. 
The  proposed method relies on a representation derived from a pre-trained convolutional neural network. Through our study, we aim to bring some light into: \textit{i)} the importance of just the visual information for solving the AP tasks (age and gender), and \textit{ii)} how complementary are both textual and visual information for the age and gender identification problems. 

The main contributions of this paper are as follows:
\begin{itemize}
	\item We built an extended \emph{multimodal} version of the PAN 2014 AP corpus, a reference data set for AP in social media. For this, we incorporated all of the images from the users' profiles contained in the original corpus.  
    \item We propose a method for AP from images based on state-of-the-art (CNNs) representation learning techniques, which have not been previously used for this task.
	\item We propose a methodology for addressing the age identification problem using posted images in Twitter. To the best of our knowledge, this is the first effort in approaching this task by using purely visual information. 
    \item We provide a comparative analysis on the importance of using textual and visual information for age and gender identification.   
    \item We evaluate the usefulness of images in the AP task, whether they are tweeted or retweeted by the users.
\end{itemize} 

The remainder of this paper is organized as follows. Section~\ref{relwork} reviews related work on AP using textual, visual and multimodal approaches. Section~\ref{visadop} describes the proposed methods for AP using images. Section~\ref{dataset} describes the adopted methodology for building a multimodal  corpus for AP in Twitter. Section~\ref{expres} presents our experimental results, which comprise quantitative and qualitative evaluation. Finally, Section~\ref{conclus} outlines our conclusions and future work directions.

\begin{table*}[!t]
\centering
\caption{State-of-art methods applied to author profiling.}
\label{tab:relwor}
\tiny\begin{tabular}{r|c|c|c|c|c|}
\cline{2-6}
\multicolumn{1}{l|}{} & \multicolumn{5}{c|}{\textbf{Author Profiling subtasks}} \\ \hline
\multicolumn{1}{|l|}{\textbf{Approaches}} & \textit{Gender} & \textit{Age} & \textit{Personality} & \textit{Interests} & \textit{Others} \\ \hline
\multicolumn{1}{|r|}{\textit{Textual}} & \cite{argamon2003gender,argamon2007mining,argamon2009automatically,burger2011discriminating,cheng2011author,goswami2009stylometric,herring2006gender,koppel2002automatically,Lopez-Monroy15,mukherjee2010improving,Ortega16,otterbacher2010inferring,peersman2011predicting,rao2010classifying,sarawgi2011gender,schler2006effects,yan2006gender} & \cite{argamon2003gender,argamon2007mining,argamon2009automatically,goswami2009stylometric,Lopez-Monroy15,nguyen2013old,nguyen2011author,Ortega16,peersman2011predicting,rao2010classifying,schler2006effects} & \cite{Litvinova16,Litvinova15} & \cite{Li2015,penas13} & sentiment anal.~\cite{Rosso16} \\ \hline
\multicolumn{1}{|r|}{\textit{Visual}} & \cite{Azam16,Hum11,Ma14,Shigenaka16,You14} & --- & \cite{Cristani13,Eftekhar14,Lovato14,Wu15} & \cite{YangHE15,You16} & retweet predict.~\cite{Can13} \\ \hline
\multicolumn{1}{|r|}{\textit{Multimodal}} & \cite{Merler15,taniguchi} & --- & --- & --- & sentiment anal.~\cite{You13} \\ \hline
\end{tabular}
\end{table*}

\section{Related Work}\label{relwork}
According to the literature, AP in social media has two main subtasks: age and gender detection (see \cite{argamon2003gender,argamon2007mining,argamon2009automatically,burger2011discriminating,cheng2011author,goswami2009stylometric,herring2006gender,koppel2002automatically,Lopez-Monroy15,mukherjee2010improving,Ortega16,otterbacher2010inferring,peersman2011predicting,rao2010classifying,sarawgi2011gender,schler2006effects,yan2006gender} for gender, and \cite{argamon2003gender,argamon2007mining,argamon2009automatically,goswami2009stylometric,Lopez-Monroy15,nguyen2013old,nguyen2011author,Ortega16,peersman2011predicting,rao2010classifying,schler2006effects} for age). Related tasks include personality prediction~\cite{Litvinova16,Litvinova15}, interests identification~\cite{Li2015,penas13} (a.k.a. genres), sentiment and emotion recognition~\cite{Rosso16}, among others. Generally speaking,  AP has been approached as a single-label classification problem, where the target profiles (e.g., males \textit{vs.} females) stand for the target classes.  

To accurately model target profiles it is necessary to extract general demographic-features that apply to heterogeneous groups of authors, and indicate, to some extent, how they use them given their native language, genre, age, etc.~\cite{argamon2003gender}. Thus, the AP task in social media is particularly challenging given the nature of Internet interactions and the informality of written language. 

Table~\ref{tab:relwor} provides a summary of related work on AP. We can distinguish three broad approaches for addressing the AP task: textual, visual and multimodal. Notice that both visual and multimodal approaches have been less studied (see last two rows in Table~\ref{tab:relwor}). Regarding the textual approach, authors have proposed combinations of textual attributes ranging from lexical features (e.g., content words~\cite{argamon2003gender} and function words~\cite{koppel2002automatically}) to syntactical features (e.g., POS-based features~\cite{bergsma2012stylometric}, personal phrases~\cite{Ortega16}, and probabilistic context-free grammars \cite{sarawgi2011gender}). Concerning the visual approach, most of the research had focused on gender recognition~\cite{Azam16,Ma14,Shigenaka16,You14}, and have only considered some general statistics about the shared images as features~\cite{Hum11}. Similarly, some works have considered visual information for the task of personality prediction~\cite{Cristani13,Eftekhar14,Lovato14,Wu15}. For instance, in \cite{Lovato14}, authors try to determine users' behavioral biometric traits analyzing their favorite images in Flickr. As features, authors considered the average size of the regions, colorfulness, and wavelet textures among others. They conclude that images are very good elements for determining people's behavior. Similarly, in \cite{Cristani13} authors proposed a method for identifying personality traits from Flickr users. Their obtained results showed a strong correlation between the personality and the favorite image of users.

Regarding multimodal approaches, it is worth mentioning that this type of strategies has just recently starting to gain attention. For example,  in \cite{taniguchi} authors proposed a weighted text-image scheme for gender classification of twitter users. The basic idea  consist on identifying a set of image categories associated for male and female classes. This process is done through the use of a CNN, which is used for determining a score for every user's image. At the end, computed scores are averaged and combined with textual information for approaching the gender identification problem. 

Contrary to previous research, this paper focuses in exploiting the visual modality to perform both age and gender identification in social media. Accordingly, we have extended a well-known benchmark corpus by means of incorporating images into users' profiles. Thus, we are able to perform a comparative analysis on the importance of using textual and visual information for the age and gender identification problem. In addition, we also evaluate the utility of images when they are original images (i.e., tweeted by the user) or reused images (i.e., retweeted from other user's post). Last, but not least, our visual approach to AP is based on state of the art visual descriptors (CNN-based). 

\section{Author profiling from images}\label{visadop}
In this section we describe the proposed visual approach for AP. Firstly, Section~\ref{representation} describes the adopted image representation. Then, Section~\ref{methods} introduces two different methods for performing AP in Twitter using information from posted images.

\subsection{CNN-based image representation}\label{representation}
Defining robust and discriminative features for images is not a trivial task. A direct way for defining these features is manually, through handcrafted features provided by experts or using a mechanical turk approach~\cite{Sorokin08}. As it is possible to imagine, this approach is highly inviable due to the number of images that would have to be labeled in a social media domain. Hence, we adopted an alternative solution based on feature learning by using a pre-trained deep learning model\footnote{Features learned by a deep model on a generic and very large dataset are transferred to another task.}~\cite{Oquab14,Yosinski14}. Deep models are composed of multiple processing layers that allow to learn representations of data with multiple levels of abstraction~\cite{Lecun15}. For instance, when an image is propagated in a pre-trained deep model, it is processed layer-by-layer transforming an array of pixel values (an image) into a representation that amplifies important aspects of the input and suppress irrelevant variations for discrimination~\cite{Lecun15}. This methodology has reported outstanding results in a number of computer vision tasks. Our intuition is that this type of representation can be beneficial in solving the posed task.

As mentioned before, our dataset is composed of images from Twitter. Thus, we choose a pre-trained model general enough to cover the visual diversity of  target images. As known, a pre-trained model will perform better, under a transfer learning scenario, when the target dataset shares a similar distribution with the source dataset~\cite{transferDL}. Accordingly,  we used the 16-layer CNN-model called VGG~\cite{Simonyan14}. This model was trained on the ImageNet dataset~\cite{Russakovsky15}, a large visual database designed for visual object recognition, including classification and detection of objects and scenes.

Every image's representation is obtained by passing its raw input (pixels) through the ConvNet model using the Caffe Library~\cite{Jia14}. Then the activation of an intermediate layer is used as the representation of the feed image. For our experiments, we choose the 4096 activations produced in the last hidden layer of the net. 

The activation of this layer produces similar values when similar images are introduced~\cite{Krizhevsky12}. Note that our CNN representation did not rely on the last layer of the net, which produces detection scores over 1000 different classes. The reason is that transferability is negatively affected by the specialization of higher layer neurons to their original task at the expense of performance on the target task~\cite{Yosinski14}. In our case, an abstract representation over 4096 neurons would be reduced to 1000 different classes. Therefore, for representing images we use the last hidden layer, and the final layer is employed only for the qualitative analysis reported in Section~\ref{analysis}.

\subsection{AP from visual and multimodal information}\label{methods}

This section describes two different visual-based methods for AP in social media. Also, it presents an effective multimodal approach that jointly uses textual and visual information, as well as a baseline method exclusively based on the use of textual information. 

\begin{enumerate}
    \item \textbf{Visual methods}. The two proposed methods for AP from images are illustrated in Figure~\ref{approach}. Both of them use the same input information and apply the same process for building the images' representation; this is, given the set of images from a user (profile), each image is passed through the pre-trained deep model obtaining its vector representation (see upper box in the figure). Next, the obtained information can be exploited in two different ways, deriving in our two proposed strategies:
    \begin{enumerate}
    	\item \textit{Individual classification}: first, each image from a user is classified individually, then, the AP class of the user is determined by means of a majority-vote strategy.
   		\item \textit{Prototype classification}: first, each user is represented by a prototype vector built by averaging the CNN representations from all his/her images. Then, this prototype is feed to a standard classier that outputs the AP class of the user.
 	\end{enumerate}

    \begin{figure}[h!tb]    
    \centering\includegraphics[width=8cm]{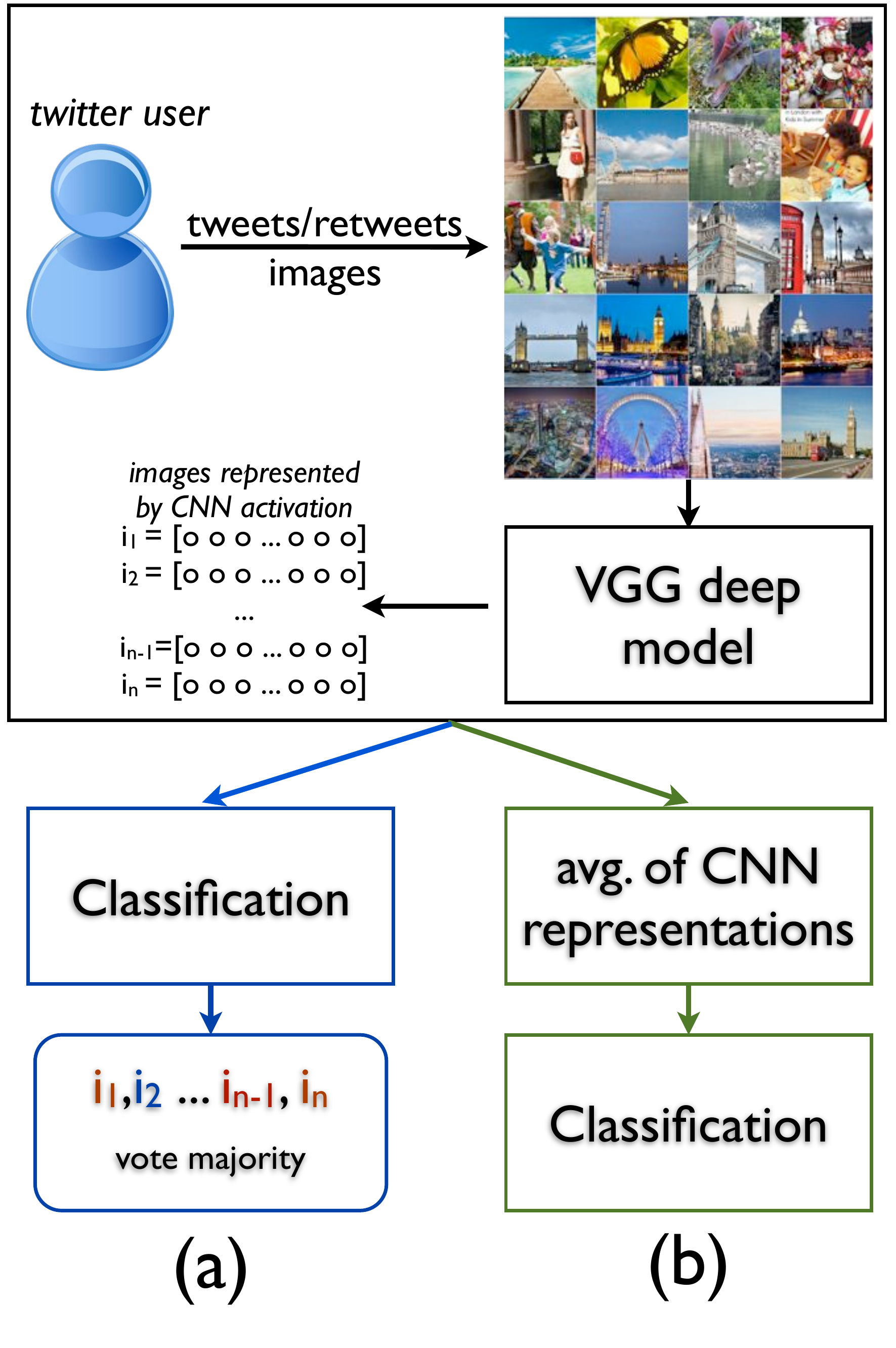}
    \caption{Two visual approaches for AP in Twitter: (a) individual-based classification and (b) prototype-based classification.}
    \label{approach}
\end{figure}
    
    \item \textbf{Multimodal method}. This method follows the same pipeline that 1.(b) for building the images' representation, however, it is called a multimodal representation since it combines visual and textual information. Specifically, we build the multimodal prototype of each user by concatenating the visual prototype with a traditional BOW representation from all the tweets that the user has posted.
    
    
    
	\item \textbf{Textual method}. As before, a prototype representation is built for each user, however, for this method we consider only textual information. Two different BOW representations were used in the experiments, containing the 2k and 10k most frequent terms from the training data respectively.
            

\end{enumerate}

More details regarding the implementation of these methods are given in Section~\ref{expres}. Next, we introduce the multimodal corpus specifically assembled for evaluating our proposed approach.

\section{A multimodal corpus for twitter AP}\label{dataset}
Images shared by social media users tend to be strongly correlated to their thematic interests as well as to their style preferences. Motivated by these facts we tackled the task of assembling a corpus considering text and images from twitter users. Mainly, we extended the PAN-2014~\cite{rangel2014overview} dataset by harvesting images from the already existing twitter users.

The PAN-2014 twitter dataset considers gender profiles (female vs. male), and five non-overlapping age profiles. It includes tweets (only textual information) from English and Spanish users. Based on this dataset we harvested more than 150,000 images, corresponding to a subset of 450 user profiles, 279 profiles in English  and 171 in Spanish\footnote{Note that the PAN-2014 corpus includes more profiles in both languages, however, for some twitter users it was impossible to download their associated images.}. The images associated to all of the users were downloaded and lined to existing user profiles, resulting in a new multimodal twitter corpus for the AP task. Next subsections present detail information from the new corpus.

\subsection{Statistics of the multimodal twitter corpus} 
Table~\ref{tab:stats} presents general statistics of the new multimodal AP corpus, which includes around 85,000 images for the English users and approximately 73,000 for the Spanish profiles. Given our interest in studying the discriminative capabilities of tweeted and retweeted images, we have separated both kinds of images; approximately 50\% of the collected images correspond to each kind. It is worth noting that although there is a considerable amount of images per profile, there is a high standard deviation in both corpora, indicating the presence of some users with very few images in their profiles. Table~\ref{tab:stats} also shows that users from the Spanish corpus posted more images (40\% more) than the users from the English corpus.

\begin{table}[!h]
\centering
\caption{General statistics of the images from the English (EN) and Spanish (SP) corpora.}
\label{tab:stats}
\begin{tabular}{cl|c||c|}
\cline{3-4}
\multicolumn{1}{l}{} &  & \textbf{EN} & \textbf{SP} \\ \hline
\multicolumn{2}{|r|}{\textbf{\# Profiles used}} & \multicolumn{1}{c||}{279} & \multicolumn{1}{c|}{171} \\ \hline
\multicolumn{2}{|r|}{\textbf{Images tweeted}} & \multicolumn{1}{c||}{44,376} & \multicolumn{1}{c|}{35,583} \\ \hline
\multicolumn{2}{|r|}{\textbf{Images retweeted}} & \multicolumn{1}{c||}{40,361} & \multicolumn{1}{c|}{37,625} \\ \hline
\multicolumn{1}{|c|}{\multirow{3}{*}{\textbf{\begin{tabular}[c]{@{}c@{}}Average\\ images ($\sigma$)\end{tabular}}}} & by \textit{profile} & 304 ($\pm$340) & 428 ($\pm$409)\\ \cline{2-4} 
\multicolumn{1}{|c|}{} & in \textit{tweet} set & 159 ($\pm$239) & 208 ($\pm$304) \\ \cline{2-4} 
\multicolumn{1}{|c|}{} & in \textit{retweet} set & 144 ($\pm$188) & 220 ($\pm$223) \\ \hline
\end{tabular}
\end{table}

Tables~\ref{tab:stages} and~\ref{tab:stgenre} present additional  statistics on the values that both variables, gender and age can take, respectively. On the one hand, Table~\ref{tab:stages} divides profiles by age ranges, i.e., 18-24, 25-34, 35-49, 50-64 and 65-N. Both languages show an important level of imbalance, being the 35-49 class the one having the greatest number of users, while extreme ages are the ones with the lowest. Nonetheless, the users from the 65-N range are the ones with the greatest number of posted images as well as the lower standard deviation values. It is also important to notice that, in both corpora, the users belonging to the 50-64 range share in average a lot of images, but show a large standard deviation, indicating the presence of some users with too many and very few images.

\begin{table}[!h]
\centering
\caption{Statistics of images shared by each age category, in both English (EN) and Spanish (SP) corpora.}
\label{tab:stages}
\scriptsize\begin{tabular}{clc|c|c|c|}
\cline{4-6}
\multicolumn{1}{l}{} &  & \multicolumn{1}{l|}{} & \multicolumn{3}{c|}{\textbf{Average images ($\sigma$)}} \\ \cline{2-6} 
\multicolumn{1}{l|}{\textit{}} & \multicolumn{1}{c|}{\textit{ages}} & \textbf{\#} & \textit{by profile} & \textit{in tweets} & \textit{in retweets} \\ \hline
\multicolumn{1}{|c|}{\multirow{5}{*}{\textbf{EN}}} & \multicolumn{1}{c|}{18-24} & 17 & 246 ($\pm$80) & 141 ($\pm$50) & 105 ($\pm$35) \\ \cline{2-6} 
\multicolumn{1}{|c|}{} & \multicolumn{1}{c|}{25-34} & 78 & 286 ($\pm$202) & 148 ($\pm$118) & 137 ($\pm$109) \\ \cline{2-6} 
\multicolumn{1}{|c|}{} & \multicolumn{1}{c|}{35-49} & 123 & 301 ($\pm$253) & 154 ($\pm$155) & 147 ($\pm$138) \\ \cline{2-6} 
\multicolumn{1}{|c|}{} & \multicolumn{1}{c|}{50-64} & 54 & 334 ($\pm$238) & 174 ($\pm$168) & 160 ($\pm$120) \\ \cline{2-6} 
\multicolumn{1}{|c|}{} & \multicolumn{1}{c|}{65-N} & 7 & 441 ($\pm$102) & 291 ($\pm$76) & 150 ($\pm$53) \\ \hline\hline
\multicolumn{1}{|c|}{\multirow{5}{*}{\textbf{SP}}} & \multicolumn{1}{l|}{18--24} & 12 & 254 ($\pm$99) & 123 ($\pm$58) & 131 ($\pm$45) \\ \cline{2-6} 
\multicolumn{1}{|c|}{} & \multicolumn{1}{l|}{25--34} & 36 & 331 ($\pm$198) & 154 ($\pm$101) & 177 ($\pm$109) \\ \cline{2-6} 
\multicolumn{1}{|c|}{} & \multicolumn{1}{l|}{35--49} & 85 & 414 ($\pm$341) & 207 ($\pm$234) & 207 ($\pm$170) \\ \cline{2-6} 
\multicolumn{1}{|c|}{} & \multicolumn{1}{l|}{50--64} & 32 & 565 ($\pm$308) & 258 ($\pm$197) & 307 ($\pm$179) \\ \cline{2-6} 
\multicolumn{1}{|c|}{} & \multicolumn{1}{l|}{65--N} & 6 & 808 ($\pm$173) & 440 ($\pm$116) & 368 ($\pm$87) \\ \hline
\end{tabular}
\end{table}

On the other hand, Table~\ref{tab:stgenre} reports some statistics for each gender profile. It is observed a balanced number of male and females users in both corpora as well as a similar number of shared images.

\begin{table}[!h]
\centering
\caption{Statistics of images shared by each gender category, in both English (EN) and Spanish (SP) corpora.}
\label{tab:stgenre}
\scriptsize\begin{tabular}{clc|c|c|c|}
\cline{4-6}
\multicolumn{1}{l}{} &  & \multicolumn{1}{l|}{} & \multicolumn{3}{c|}{\textbf{Average images ($\sigma$)}} \\ \cline{2-6} 
\multicolumn{1}{l|}{\textit{}} & \multicolumn{1}{l|}{\textit{gender}} & \textbf{\#} & \textit{by profile} & \textit{in tweets} & \textit{in retweets} \\ \hline
\multicolumn{1}{|c|}{\multirow{2}{*}{\textbf{EN}}} & \multicolumn{1}{c|}{F $\Venus$} & 140 & 162 ($\pm$294) & 83 ($\pm$182) & 79 ($\pm$158) \\ \cline{2-6} 
\multicolumn{1}{|c|}{} & \multicolumn{1}{c|}{M $\Mars$} & 139 & 141 ($\pm$274) & 75 ($\pm$192) & 65 ($\pm$144) \\ \hline\hline
\multicolumn{1}{|c|}{\multirow{2}{*}{\textbf{SP}}} & \multicolumn{1}{c|}{F $\Venus$} & 86 & 228 ($\pm$372) & 104 ($\pm$236) & 124 ($\pm$205) \\ \cline{2-6} 
\multicolumn{1}{|c|}{} & \multicolumn{1}{c|}{M $\Mars$} & 85 & 200 ($\pm$347) & 104 ($\pm$242) & 96 ($\pm$178) \\ \hline
\end{tabular}
\end{table}

\section{Experimental results}\label{expres}
This section presents experimental results on the multimodal twitter corpus introduced previously. This section is divided in two parts: (1) a comparison among different methods for performing AP, followed by (2) a discussion based on a purely visual evaluation. 
Overall, our aim is to show how useful the images are to approach the AP task.

\begin{table*}[!hb]
\centering
\caption{Comparison among methods based on textual, visual and multimodal approaches for performing author profiling task.}
\label{tab:compare}
\scriptsize\begin{tabular}{cl|c|c||c|c|}
\cline{3-6}
\multicolumn{1}{l}{} &  & \multicolumn{4}{c|}{\textit{evaluating}} \\ \hline
\multicolumn{1}{|c|}{\textit{approach}} & \textit{methods} & age (EN) & gender (EN) & age (SP) & \multicolumn{1}{l|}{gender (SP)} \\ \hline
\multicolumn{1}{|c|}{\multirow{2}{*}{Textual}} & T1: BoW (2k) & 0.394 & 0.741 & 0.481 & 0.601 \\ \cline{2-6} 
\multicolumn{1}{|c|}{} & T2: BoW (10k) & 0.409 & 0.755 & \textbf{0.505} & \textbf{0.703} \\ \hline\hline
\multicolumn{1}{|c|}{\multirow{2}{*}{Visual}} 
& V3: LL-CNN (all-imgs) & 0.349 & 0.526 & 0.481 & 0.524 \\ \cline{2-6} 
\multicolumn{1}{|c|}{} & V4: LL-CNN AVG (all-imgs) & 0.390 & 0.700 & 0.380 & 0.650 \\ \hline\hline
\multicolumn{1}{|c|}{\multirow{2}{*}{Multimodal}} 
& M3: T1+V4 & 0.414 & 0.775 & 0.451 & 0.685 \\ \cline{2-6} 
\multicolumn{1}{|c|}{} & M6: T2+V4 & \textbf{0.423} & \textbf{0.778} & 0.433 & 0.642 \\ \hline
\end{tabular}
\end{table*}

\subsection{Comparison of textual, visual and multimodal methods for AP}
This subsection compares the performance (classification accuracy) of the methods introduced in Section~\ref{visadop} when approaching the AP task. Evaluation is carried out by profile, allowing us to make a fair comparison among the different approaches. In order to provide comparable results, using the profile ID in the PAN-2014 twitter corpus, we construct 10 subject-independent partitions (including at least one subject from each class in each partition) and we adopt 10-fold cross-validation strategy for evaluation. As expected, partitions are unbalanced with respect to number of images. We used a SVM using LibLinear~\cite{Fan08} for classification, making a direct comparison among evaluated approaches. Thereby, although different representations are used by the methods, the information came from the same profiles. Besides, the evaluated representations used by the methods do not include any preprocessing, this is to perform a fair comparison among different modalities. 

Considering the proposed visual methods, four variants were evaluated: three of them using individual classification, considering all-images (i.e. V3 in Table~\ref{tab:compare}); and one using prototype classification in the all-images subset (V4 in Table~\ref{tab:compare}). In this last case, we decided to include only the variant with the biggest performance.

Table~\ref{tab:compare} presents the obtained results, highlighting in bold  the best obtained result by column. On the one hand, surprisingly, the evaluation in English corpus reveals that using a multimodal approach is better for detecting age and gender, when all images are collapsed into a prototype. Regarding the visual methods, they perform poorly when compared to the textual one (except V4).

On the other hand, the best results on the evaluation in the Spanish corpus are achieved by textual approaches, especially when more terms in BoW are considered, i.e. 10k. However, it is worth noticing that for gender identification, the V4 method is very competitive (as in the case of the English corpus). The multimodal approaches obtained quite competitive performance as well, but they were not able to outperform the text-only results.

With the goal of studying the discriminative properties of tweeted and retweeted images, we performed an additional experiment in which both types of images were separated and then we evaluated our proposed visual and multimodal approaches. The obtained results reveals an accuracy performance of 0.432 for the V3 method using retweeted images, surpassing the results obtained in age identification for English corpus. However, none other improvements were observed.

\subsection{How much a single image says about the user}
This scenario aims to evaluate globally the visual information in images. In this setting, every image from a profile is an individual instance where the label is the same that was assigned to the profile. We have evaluated gender and age for the two corpora in PAN-2014. For the sake of the evaluation, we have included the probability for each class evaluated. 

From Table~\ref{tab:eval0}, we can confirm the usefulness of visual information for performing gender identification, obtaining in most of the cases better results than the class probability in both corpora. Besides, a better performance is observed in the female gender. Instead, the age recognition task seems to be more challenging under this scenario, it seems that images posted by users are more diverse in terms of age. 

Interestingly, although, the  highest performances are reached for the majority classes, these are only higher than the class probability for the Spanish corpus. Whereas for the English language, results for three age intervals surpassed the class probability.

\begin{table}[!h]
\centering
\caption{Accuracy performance on age and gender identification.}
\label{tab:eval0}
\begin{tabular}{clll}
\multicolumn{1}{l}{} &  &  &  \\ \cline{2-4} 
\multicolumn{1}{l|}{} & \multicolumn{1}{l|}{\textit{corpus}$\rightarrow$} & \multicolumn{1}{l||}{English [P*]} & \multicolumn{1}{l|}{Spanish [P*]} \\ \hline
\multicolumn{1}{|c|}{\multirow{5}{*}{\textbf{\begin{tabular}[c]{@{}c@{}}Age \\ ranges\end{tabular}}}} & \multicolumn{1}{c|}{18-24} & \multicolumn{1}{c||}{0.071 [0.049]} & \multicolumn{1}{c|}{0.040 [0.042]} \\ \cline{2-4} 
\multicolumn{1}{|c|}{} & \multicolumn{1}{c|}{25-34} & \multicolumn{1}{c||}{0.269 [0.264]} & \multicolumn{1}{c|}{0.080 [0.163]} \\ \cline{2-4} 
\multicolumn{1}{|c|}{} & \multicolumn{1}{c|}{35-49} & \multicolumn{1}{c||}{0.333 [0.437]} & \multicolumn{1}{c|}{0.580 [0.482]} \\ \cline{2-4} 
\multicolumn{1}{|c|}{} & \multicolumn{1}{c|}{50-64} & \multicolumn{1}{c||}{0.343 [0.213]} & \multicolumn{1}{c|}{0.220 [0.247]} \\ \cline{2-4} 
\multicolumn{1}{|c|}{} & \multicolumn{1}{c|}{65-N} & \multicolumn{1}{c||}{0.010 [0.036]} & \multicolumn{1}{c|}{0.060 [0.066]} \\ \cline{2-4} 
\multicolumn{1}{|c|}{} & \multicolumn{1}{c|}{\textit{accuracy}} & \multicolumn{1}{c||}{0.296} & \multicolumn{1}{c|}{0.355} \\ \hline
\multicolumn{1}{|l|}{\multirow{2}{*}{\textbf{Gender}}} & \multicolumn{1}{c|}{Female $\Venus$} & \multicolumn{1}{c||}{0.578 [0.535]} & \multicolumn{1}{c|}{0.548 [0.467]} \\ \cline{2-4} 
\multicolumn{1}{|l|}{} & \multicolumn{1}{c|}{Male $\Mars$} & \multicolumn{1}{c||}{0.503 [0.465]} & \multicolumn{1}{c|}{0.509 [0.532]} \\ \cline{2-4} 
\multicolumn{1}{|l|}{} & \multicolumn{1}{c|}{\textit{accuracy}} & \multicolumn{1}{c||}{0.546} & \multicolumn{1}{c|}{0.531} \\ \hline
\end{tabular}\\\scriptsize{$^*$Class probability.}
\end{table}

\begin{table*}[!hb]
\centering
\caption{Accuracy performance considering origin of the images.}
\label{tab:eval1}\scriptsize{
\begin{tabular}{r|c|c|c|c|c|c|}
\cline{2-7}
\multicolumn{1}{l|}{} & \multicolumn{2}{l|}{(a) \textit{\textbf{testing all-imgs/{[}training with{]}}}} & \multicolumn{2}{l|}{(b) \textit{\textbf{{[}testing with{]}/training all-imgs}}} & \multicolumn{2}{l|}{(c) \textit{\textbf{{[}testing/training{]} with}}} \\ \hline
\multicolumn{1}{|l|}{\textit{evaluating} $\downarrow$} & \textit{tweets} & \textit{retweets} & \textit{tweets} & \textit{retweets} & \textit{tweets} & \textit{retweets} \\ \hline
\multicolumn{1}{|r|}{age (EN)} & 0.288 & 0.322 & 0.298 & 0.294 & 0.263 & 0.340 \\ \hline
\multicolumn{1}{|r|}{gender (EN)} & 0.515 & 0.535 & 0.550 & 0.541 & 0.519 & 0.544 \\ \hline\hline
\multicolumn{1}{|r|}{age (SP)} & 0.350 & 0.290 & 0.357 & 0.349 & 0.357 & 0.282 \\ \hline
\multicolumn{1}{|r|}{gender (SP)} & 0.532 & 0.510 & 0.525 & 0.544 & 0.549 & 0.504 \\ \hline
\end{tabular}}
\end{table*}

\subsubsection{The importance of tweeted and retweeted images}
In view of the results obtained earlier, we repeated the experiment (same evaluation protocol), but this time separating the source of the images, i.e. tweeted and retweeted. This distinction aimed at answering the following questions: 
\begin{enumerate}
	\item Do the posted (tweet) and re-shared (retweet) images express author's interests in the same way?
    \item Do any of these ways of sharing images give more information about the user's profile?
\end{enumerate}

Thus, three specific scenarios were defined according to the source of the images: \textbf{(a)} testing all-images (from both sources) and training with one source, i.e. tweeted images, retweeted images, and all-images; \textbf{(b)} training with all-images and testing in one of the two sources; and finally \textbf{(c)} testing and training using one of the two sources. The obtained results are presented in Table~\ref{tab:eval1}.

From Table~\ref{tab:eval1} we can stress the following:
\begin{itemize}
	\item Results for scenario \textbf{(a)} indicate that only using an image source for training is comparable and even better to use both sources together, in age and gender identification (see results from scenario \textbf{(a)} when compared with results from Table~\ref{tab:eval0}). It is interesting that English and Spanish take advantage of different sources, English presents better performances using retweeted images for training, whilst Spanish from tweeted images.
    \item Scenario \textbf{(b)} allows us to compare the results obtained in each source by including all-images as training. In the case of age, only the retweeted images in Spanish have presented a considerable increment in performance, i.e. from 0.29 to 0.34. 
    \item Scenario \textbf{(c)} allows us to compare the results obtained with those obtained in scenario \textbf{(b)}. Here, the training is reduced to individual sources in comparison to scenario \textbf{(b)} that uses all-images. In general, we can see decrements in age and gender identification over English, with an exception in retweeted images where it has obtained better results. Instead, the tweeted images for gender identification in Spanish has been the only with an increment in performance.
\end{itemize}

Summarizing the obtained results from the experiments in this section, we have showed that it is feasible, using only the visual information from posted images, to approach the gender identification task and in some cases the age identification as well. Moreover, we have found that, in fact,  the image source matters (i.e. tweeted or retweeted), and it is possible to exploit it for achieving better results on the age and gender identification.

\subsection{A picture is worth a thousand words}
Inspired by the saying \textit{'a picture is worth a thousand words'}, we decided to classify individual profiles taken randomly. Some works have followed the same idea, for instance~\cite{Hum11} has performed a statistical analysis by genders, and~\cite{You16} has constructed a dataset from Pinterest with the aim to classify images from a user into categories. Instead, here explicitly each profile is defined by a tweet composed of only 1000 words, and after classified. In return, images from the same profile are classified. Both, classification schemes are compared trying to answer whether it is possible to say more with a picture than with thousand words.

In order to approximate the same conditions for both classification scenarios, the training instances are provided from the same profiles but of course using its respective representation. For the case of textual approach, samples of 1000 words are used for textual representation, repeating as many subsets of this size can be extracted from the profile. Instead, all images in profiles are classified for the visual approach. Obtained results are presented in Tables~\ref{tab:1img_age} and~\ref{tab:1img_gen} for age and gender identification, respectively.

In general, results from Table~\ref{tab:1img_age} indicate that it is possible to identify with reasonable accuracy some age ranges by  using images only, this holds for  both languages. Especially, minority classes where using only 1000 words it is not enough. Even more interesting is to observe that the image source is important, presenting better results than using all-images without making a distinction.

\begin{table}[!h]
\centering
\caption{Accuracy performance for age ranges.}
\label{tab:1img_age}
\tiny\begin{tabular}{ll|c|c|c|c|c|}
\cline{3-7}
 &  & \multicolumn{2}{c|}{\textbf{Textual}} & \multicolumn{3}{c|}{\textbf{Visual}} \\ \cline{2-7} 
\multicolumn{1}{l|}{} & \textit{\textbf{ages}} & \multicolumn{1}{l|}{\textit{BoW (2k)}} & \multicolumn{1}{l|}{\textit{BoW (10k)}} & \multicolumn{1}{l|}{\textit{all-images}} & \multicolumn{1}{l|}{\textit{tweets}} & \multicolumn{1}{l|}{\textit{retweets}} \\ \hline
\multicolumn{1}{|l|}{\multirow{5}{*}{\textbf{ES}}} & \multicolumn{1}{c|}{\textit{18-24}} & 0\% & 0\% & 4.6\% & \textbf{12.6\%} & 4\% \\ \cline{2-7} 
\multicolumn{1}{|l|}{} & \multicolumn{1}{c|}{\textit{25-34}} & 0\% & 14.2\% & \textbf{29.7\%} & 13.8\% & 17\% \\ \cline{2-7} 
\multicolumn{1}{|l|}{} & \multicolumn{1}{c|}{\textit{35-49}} & \textbf{100\%} & \textbf{100\%} & 30\% & 41.8\% & 33.1\% \\ \cline{2-7} 
\multicolumn{1}{|l|}{} & \multicolumn{1}{c|}{\textit{50-64}} & 14.2\% & 14.2\% & 38.5\% & \textbf{41.8\%} & 33.1\% \\ \cline{2-7} 
\multicolumn{1}{|l|}{} & \multicolumn{1}{c|}{\textit{65-N}} & 0\% & 0\% & 0\% & \textbf{8.3\%} & 0\% \\ \hline
\multicolumn{1}{|l|}{\multirow{5}{*}{\textbf{EN}}} & 18-24 & 0\% & 0\% & 3.1\% & 3.1\% & \textbf{14.2\%} \\ \cline{2-7} 
\multicolumn{1}{|l|}{} & 25-34 & 0\% & 11.1\% & 28.1\% & \textbf{39.6\%} & 31.4\% \\ \cline{2-7} 
\multicolumn{1}{|l|}{} & 35-49 & 33\% & \textbf{50\%} & 38.3\% & 27.6\% & 8.5\% \\ \cline{2-7} 
\multicolumn{1}{|l|}{} & 50-64 & 18.1\% & 18.1\% & 20.1\% & 29.1\% & \textbf{29.4\%} \\ \cline{2-7} 
\multicolumn{1}{|l|}{} & 65-N & 0\% & \textbf{14.2\%} & 0.3\% & 1.4\% & 9.1\% \\ \hline
\end{tabular}
\end{table}

On the other hand, results obtained by gender identification indicate that it is kind of easier to determine whether a profile belongs a female/male person by their images than by their words. Interestingly, for female gender it is better using the tweeted images for performing the identification, while for the male gender using retweeted images works better.

\begin{table}[!h]
\centering
\caption{Accuracy performance for gender.}
\label{tab:1img_gen}
\scriptsize\begin{tabular}{lc|c|c||c|c|}
\cline{3-6}
 & \multicolumn{1}{l|}{} & \multicolumn{2}{c||}{\textbf{Spanish}} & \multicolumn{2}{c|}{\textbf{English}} \\ \cline{3-6} 
 & \multicolumn{1}{r|}{\textit{\textbf{}}} & \multicolumn{1}{c|}{\textbf{F} $\Venus$} & \multicolumn{1}{c||}{\textbf{M} $\Mars$} & \multicolumn{1}{c|}{\textbf{F} $\Venus$} & \multicolumn{1}{c|}{\textbf{M} $\Mars$} \\ \hline
\multicolumn{1}{|c|}{\multirow{2}{*}{\textbf{Textual}}} & \textit{BoW (2k)} & 40\% & 0\% & 20\% & 25\% \\ \cline{2-6} 
\multicolumn{1}{|c|}{} & \textit{BoW (10k)} & 60\% & 20\% & 20\% & 50\% \\ \hline
\multicolumn{1}{|l|}{\multirow{3}{*}{\textbf{Visual}}} & \textit{all-images} & 54.6\% & 55.5\% & 52.5\% & 55.7\% \\ \cline{2-6} 
\multicolumn{1}{|l|}{} & \textit{tweets} & \textbf{75.7\%} & 37\% & \textbf{95.7\%} & 5.3\% \\ \cline{2-6} 
\multicolumn{1}{|l|}{} & \textit{retweets} & 55.2\% & \textbf{59.2\%} & 38.9\% & \textbf{62.8\%} \\ \hline
\end{tabular}
\end{table}

\begin{figure*}[!hb]
    \centering    
    \begin{subfigure}[b]{0.22\textwidth}
        \includegraphics[width=\hsize]{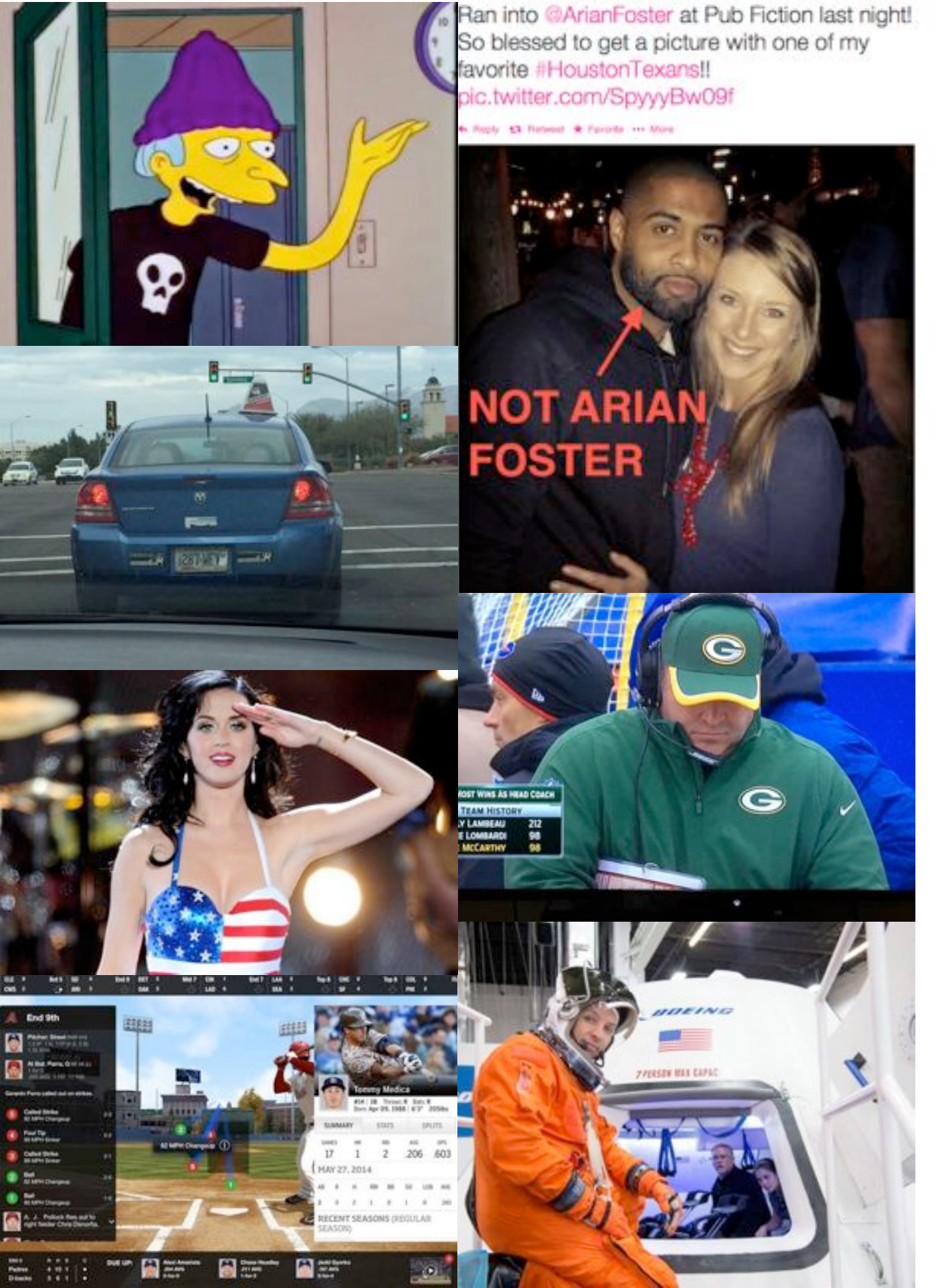}
        \caption{Male $\Mars$ (EN) images.}
        \label{fig:manen}
    \end{subfigure}
    \begin{subfigure}[b]{0.5\textwidth}
        \includegraphics[width=\hsize]{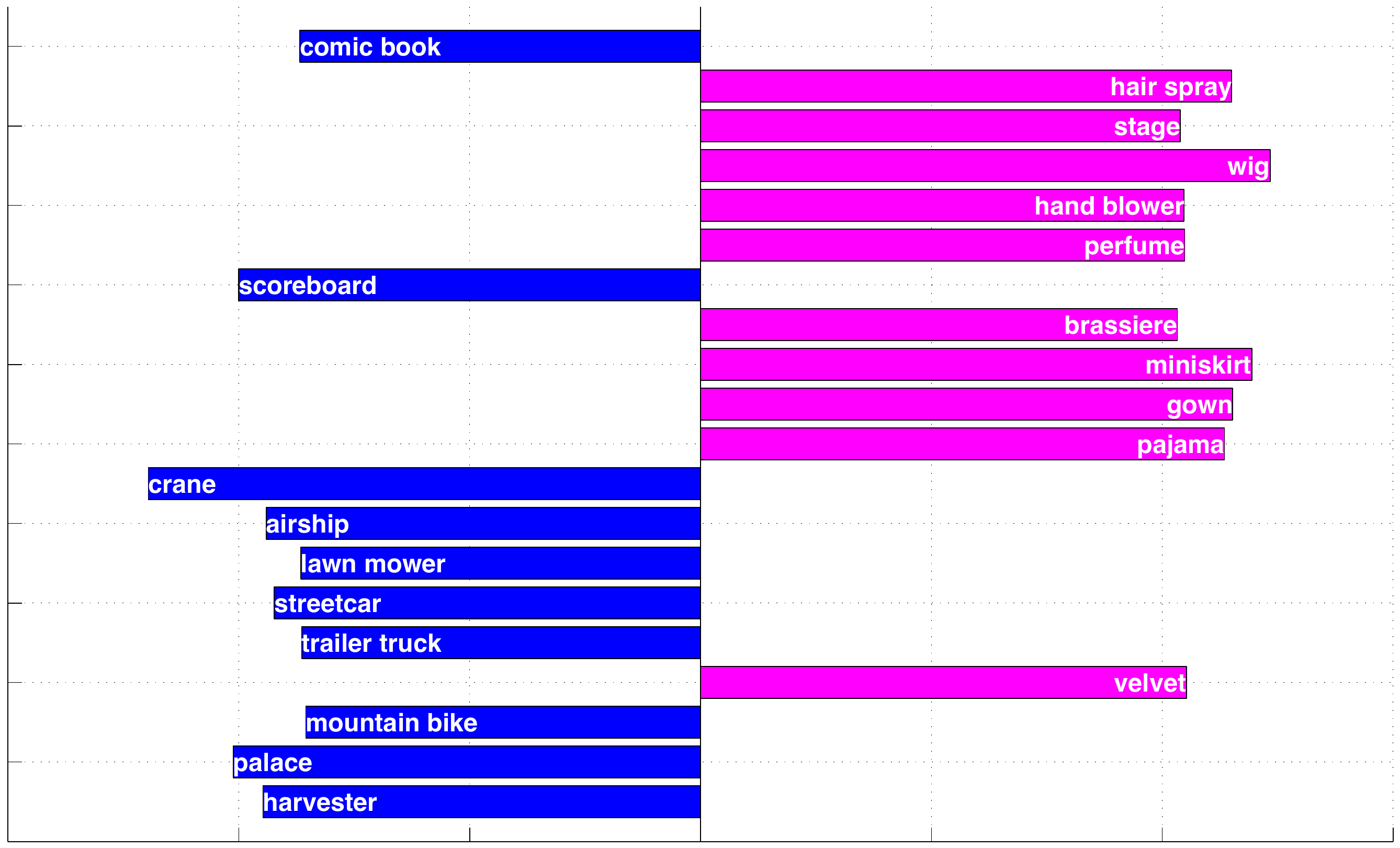}
        \caption{Differences among labels from English corpus.}
        \label{fig:genen}
    \end{subfigure}
    \begin{subfigure}[b]{0.22\textwidth}
        \includegraphics[width=\hsize]{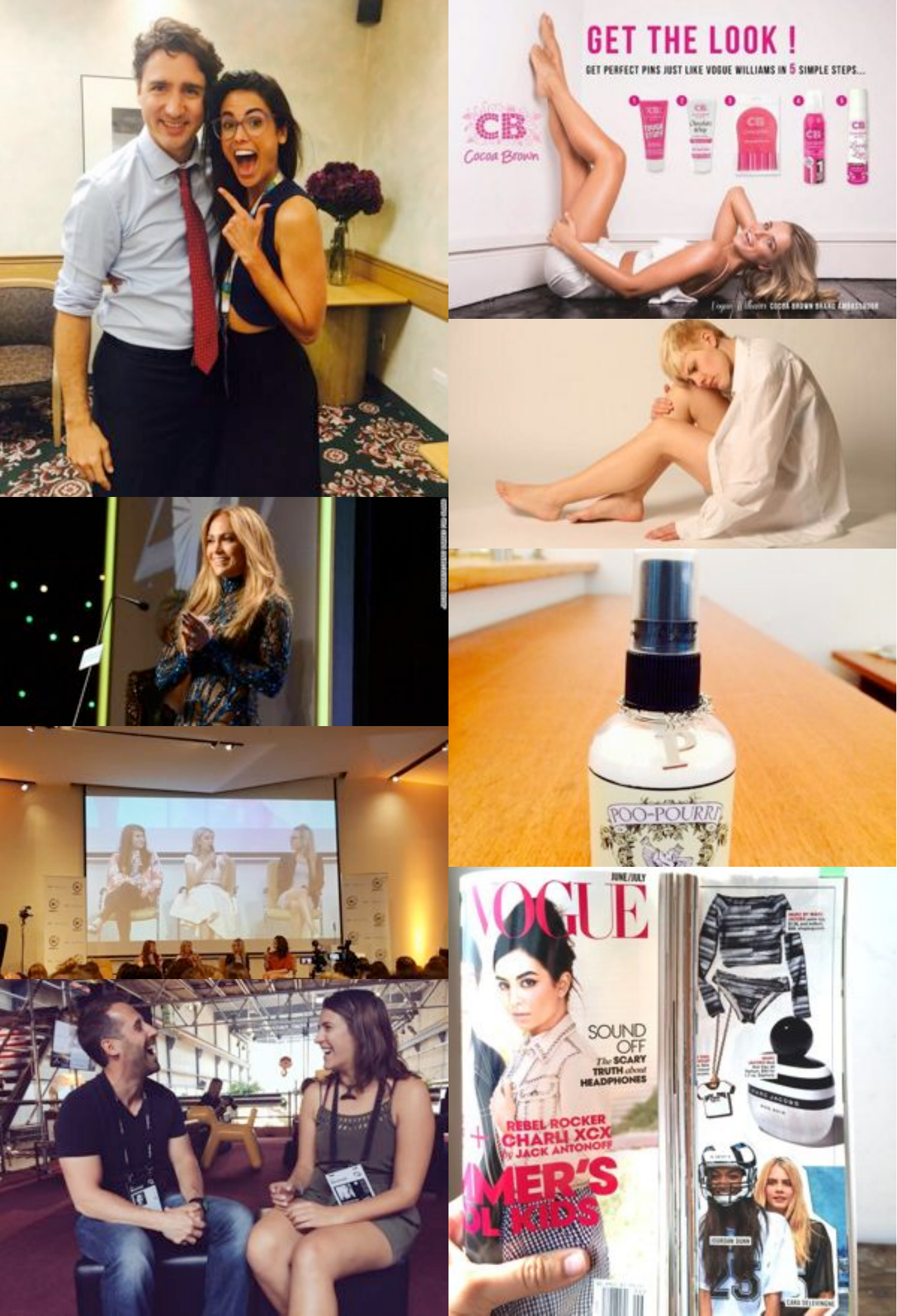}
        \caption{Female $\Venus$ (EN) images.}
        \label{fig:femen}
    \end{subfigure}\\
    \begin{subfigure}[b]{0.22\textwidth}
        \includegraphics[width=\hsize]{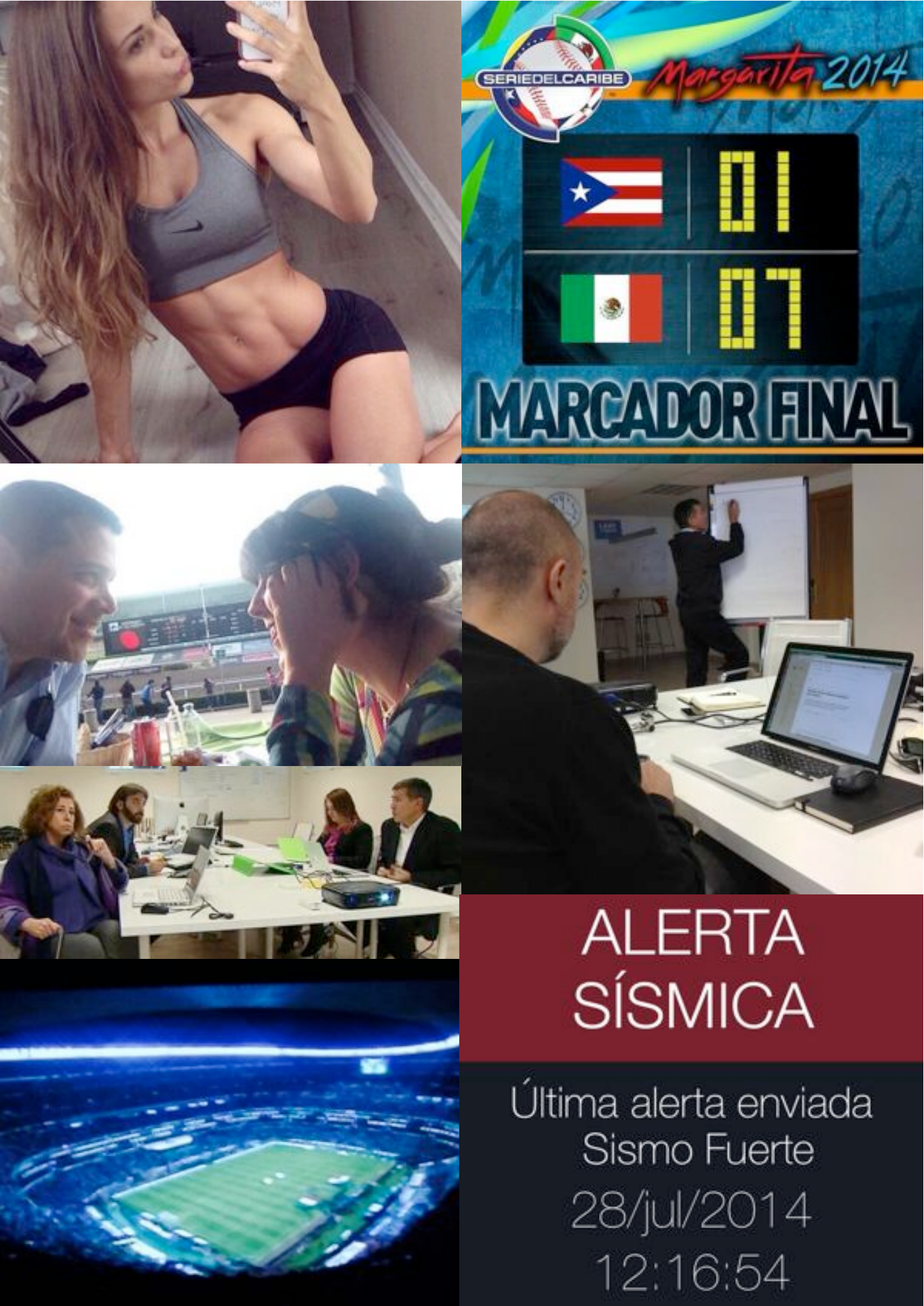}
        \caption{Male $\Mars$ (SP) images.}
        \label{fig:manes}
    \end{subfigure}
    \begin{subfigure}[b]{0.5\textwidth}
        \includegraphics[width=\hsize]{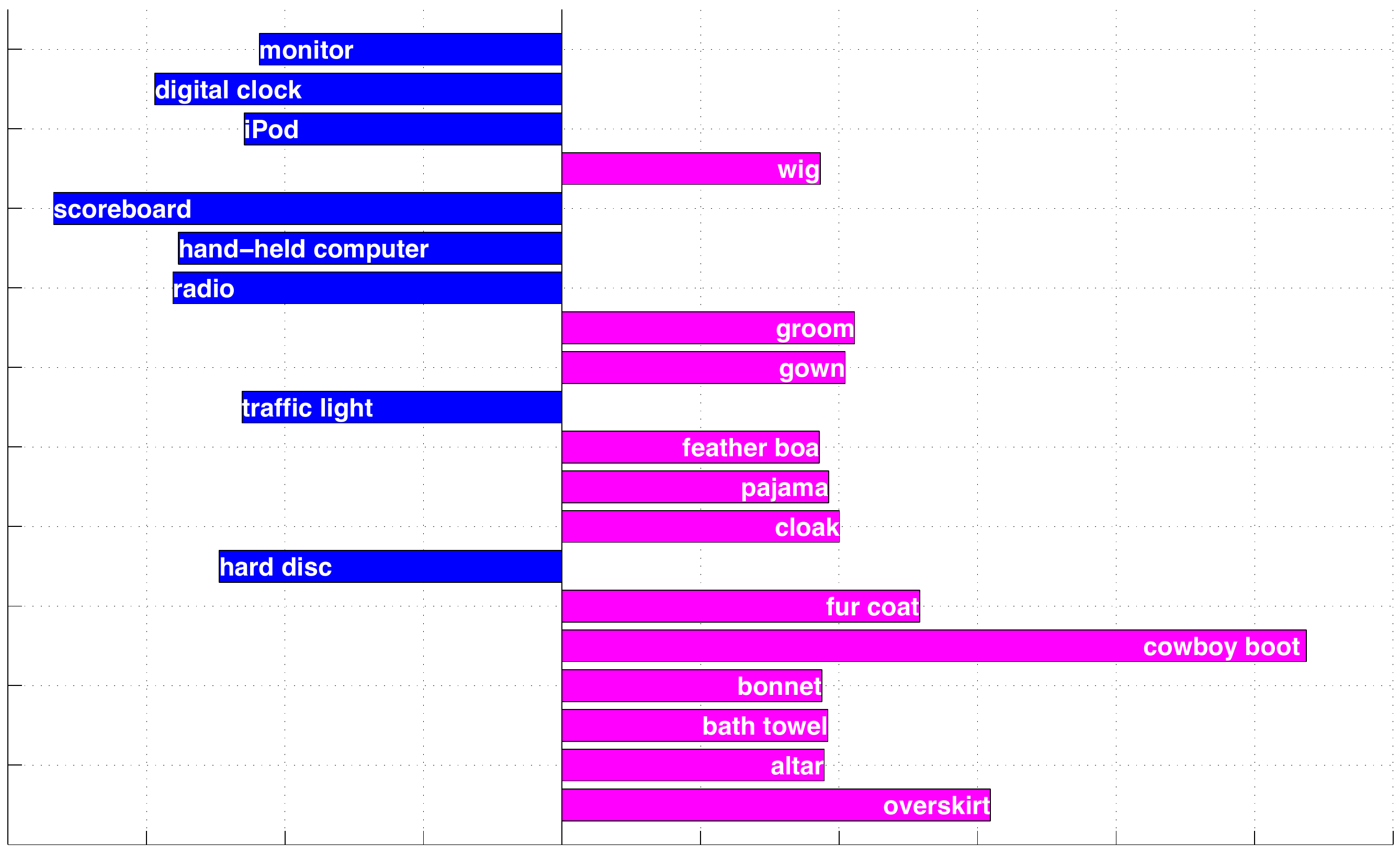}
        \caption{Differences among labels from Spanish corpus.}
        \label{fig:genes}
    \end{subfigure}
    \begin{subfigure}[b]{0.22\textwidth}
        \includegraphics[width=\hsize]{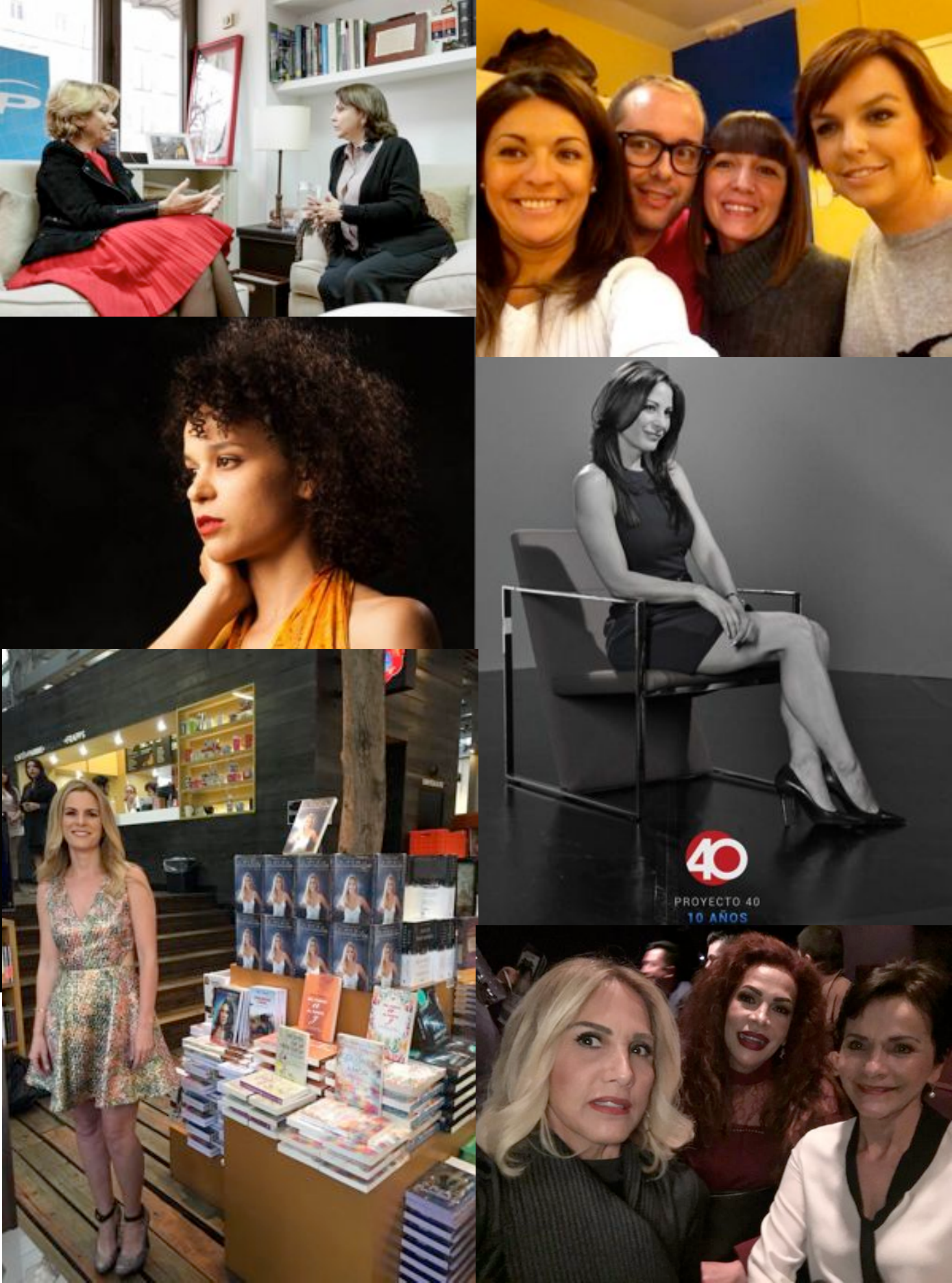}
        \caption{Female $\Venus$ (SP) images.}
        \label{fig:femes}
    \end{subfigure}
    \caption{Comparison of frequent labels found in male and female profiles.}
    \label{words_manvsfemale}
\end{figure*}

\subsection{Qualitative analysis of the posted images}\label{analysis}
This section presents qualitative experiments in order to show how useful are images for exploring information in Twitter. Our aim to perform this study is to show the visual evidence left by users in a lapse of time. 

For this experiment we represented images with the final layer of the considered CNN. The dimensions correspond to the 1000 categories in ImageNet~\cite{Russakovsky15}, thus the higher the value is, then the more likely the corresponding category appears. Of course it is unlikely that images posted in a 'wild' scenario could be represented by only 1000 classes. However, through this kind of experiments we can define user preferences using denotative descriptions, i.e. labels assigned.

Hence, in order to analyze the content in images posted by the users, we  labeled all-images using these 1,000 ImageNet categories~\cite{Simonyan14}. After classification, images in a specific gender or age range are concentrated in normalized histograms. A similar study has been carried out in~\cite{YangHE15} using Pinterest images under the \textit{Travel} category, nevertheless, the authors intended to answer whether user-generated visual contents had  predictive capabilities for users' preferences beyond labels. 

Figure~\ref{words_manvsfemale} shows a list of words with frequency order (top to down), comparing how often they are used by gender in both corpora, i.e. male versus female. Besides, a sample of images posted accompany both genders. For producing such word list, the difference is calculated over normalized histograms of genders. We have taken 20 words with the biggest difference considering equal number of words in favor of each gender.

On the one hand, in the English corpus (see \ref{fig:manen}, \ref{fig:genen} and \ref{fig:femen} in Figure~\ref{words_manvsfemale}), the male gender users seem to post images associated to topics as sports (i.e. 'mountain bike', 'scoreboard', etc.), machines and vehicles (i.e. 'airship', 'trailer truck', 'streetcar', etc.). Whereas, for female gender users, there is a topic related to beauty products ('hair spray', 'perfume', 'hand blower'), and another associated to fashion (i.e. 'velvet', 'wig', 'pajama', etc.).

On the other hand, in the Spanish corpus (see \ref{fig:manes}, \ref{fig:genes} and \ref{fig:femes} in Figure~\ref{words_manvsfemale}), the male gender users are prone to post images that include sports (i.e. 'scoreboard' and 'digital clock') as well as technology (i.e. 'monitor', 'hand-held computer', 'iPod', etc.). While the female gender users posted images related to fashion (i.e. 'wig', 'miniskirt', 'bonnet', etc.).

\begin{figure*}
    \centering    
    \begin{subfigure}[b]{0.4\textwidth}
        \includegraphics[width=\hsize]{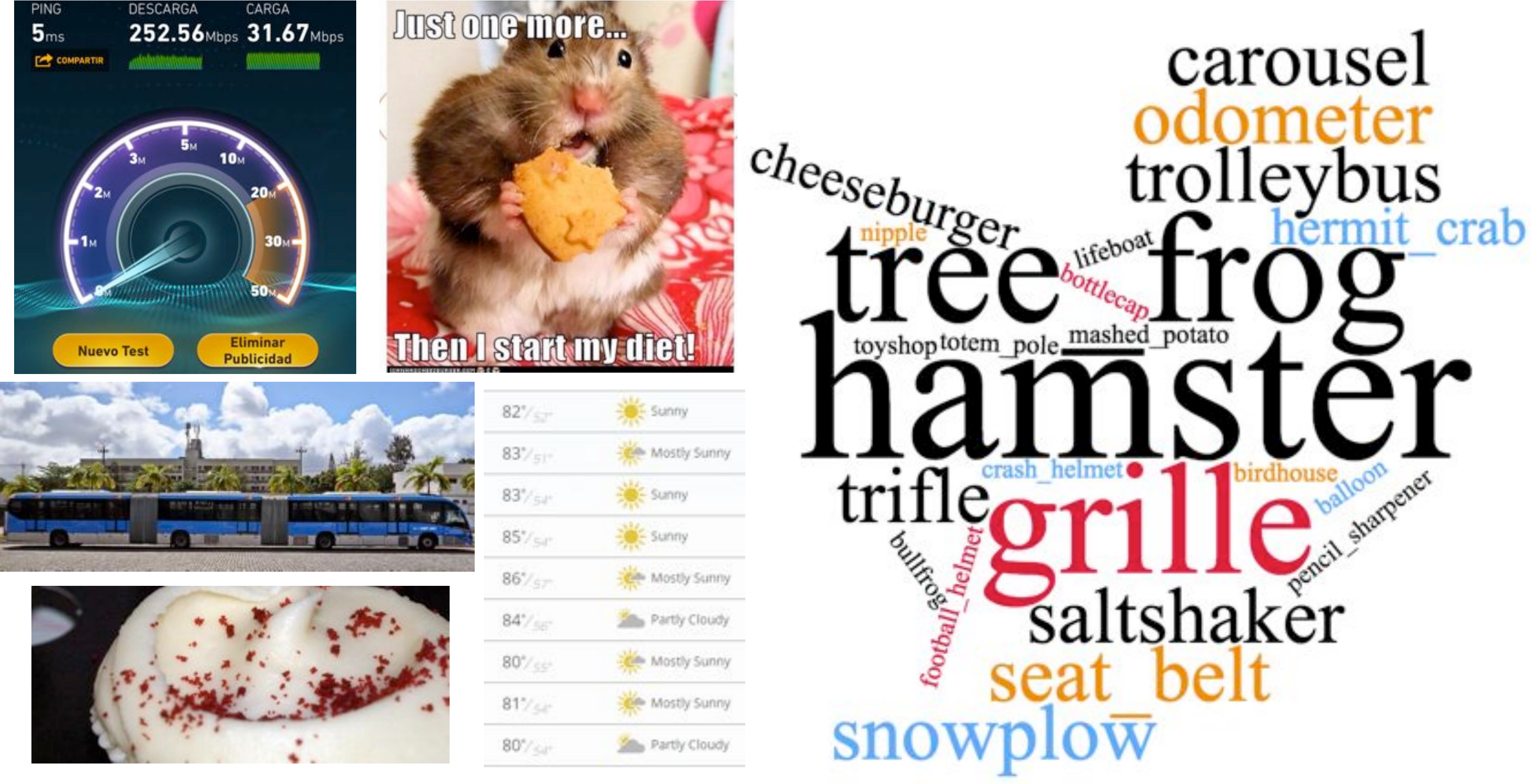}
        \caption{From 18-24 ages (English).}
        \label{fig:en18}
    \end{subfigure}
    \begin{subfigure}[b]{0.4\textwidth}
        \includegraphics[width=\hsize]{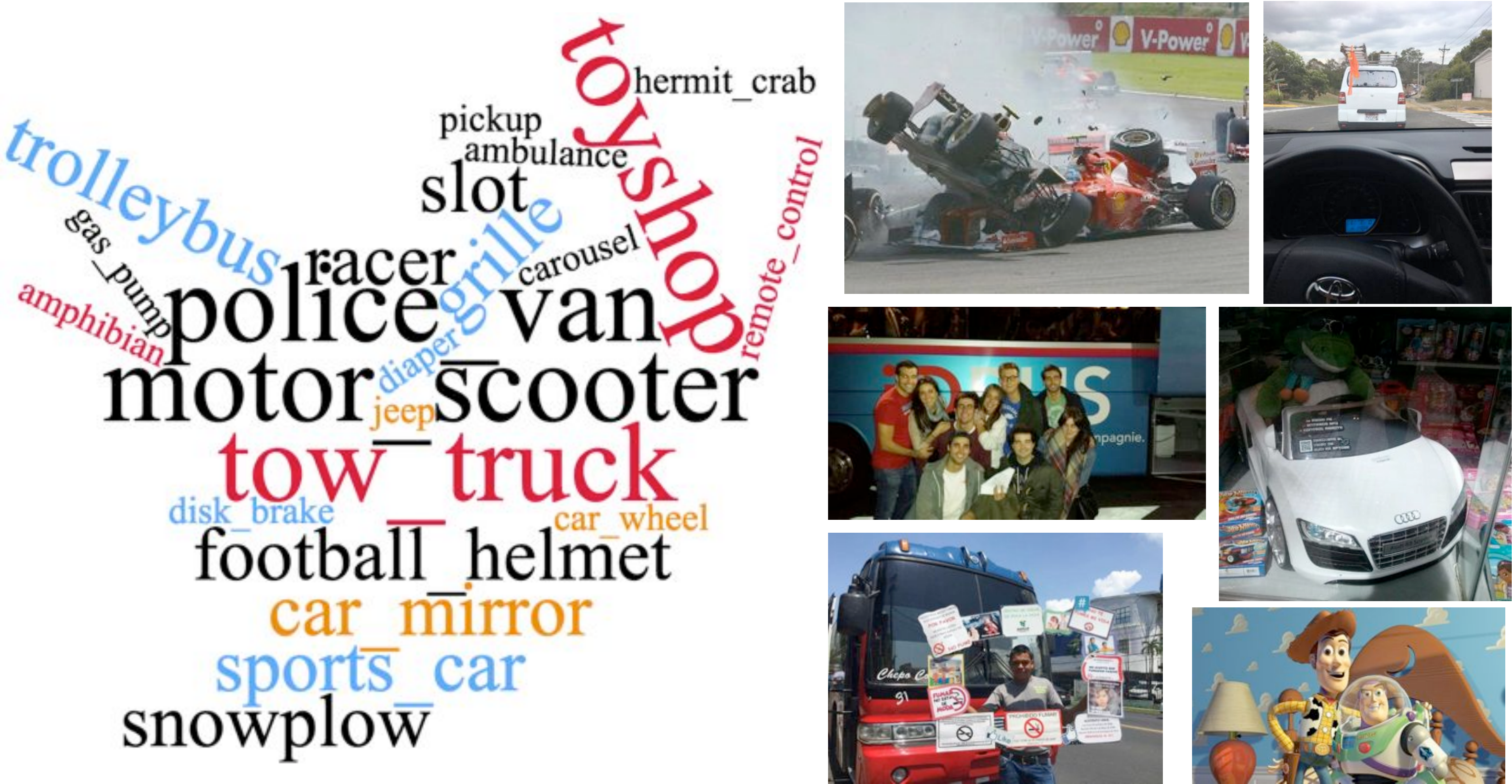}
        \caption{From 18-24 ages (Spanish).}
        \label{fig:es18}
    \end{subfigure}\\
    \begin{subfigure}[b]{0.4\textwidth}
        \includegraphics[width=\hsize]{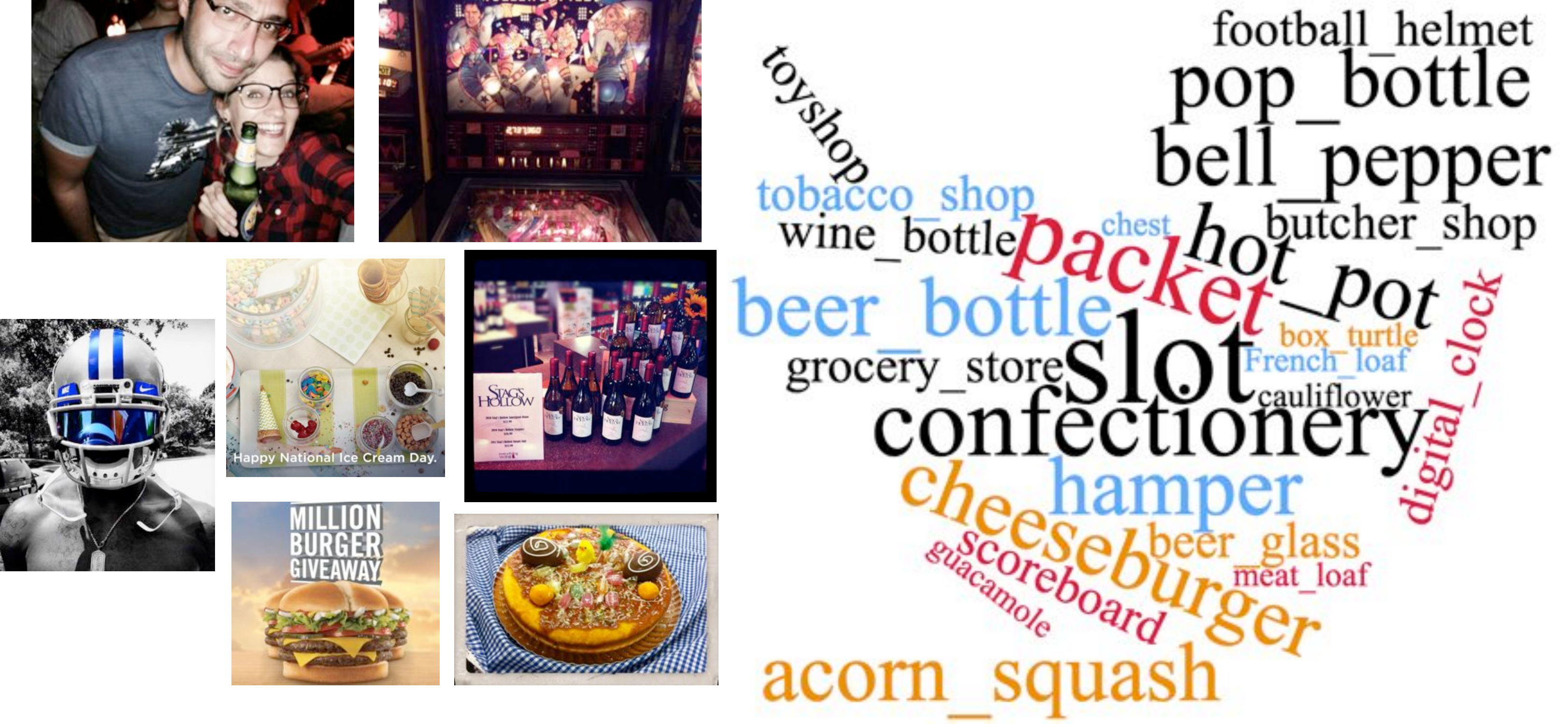}
        \caption{From 25-34 ages (English).}
        \label{fig:en25}
    \end{subfigure}
    \begin{subfigure}[b]{0.4\textwidth}
        \includegraphics[width=\hsize]{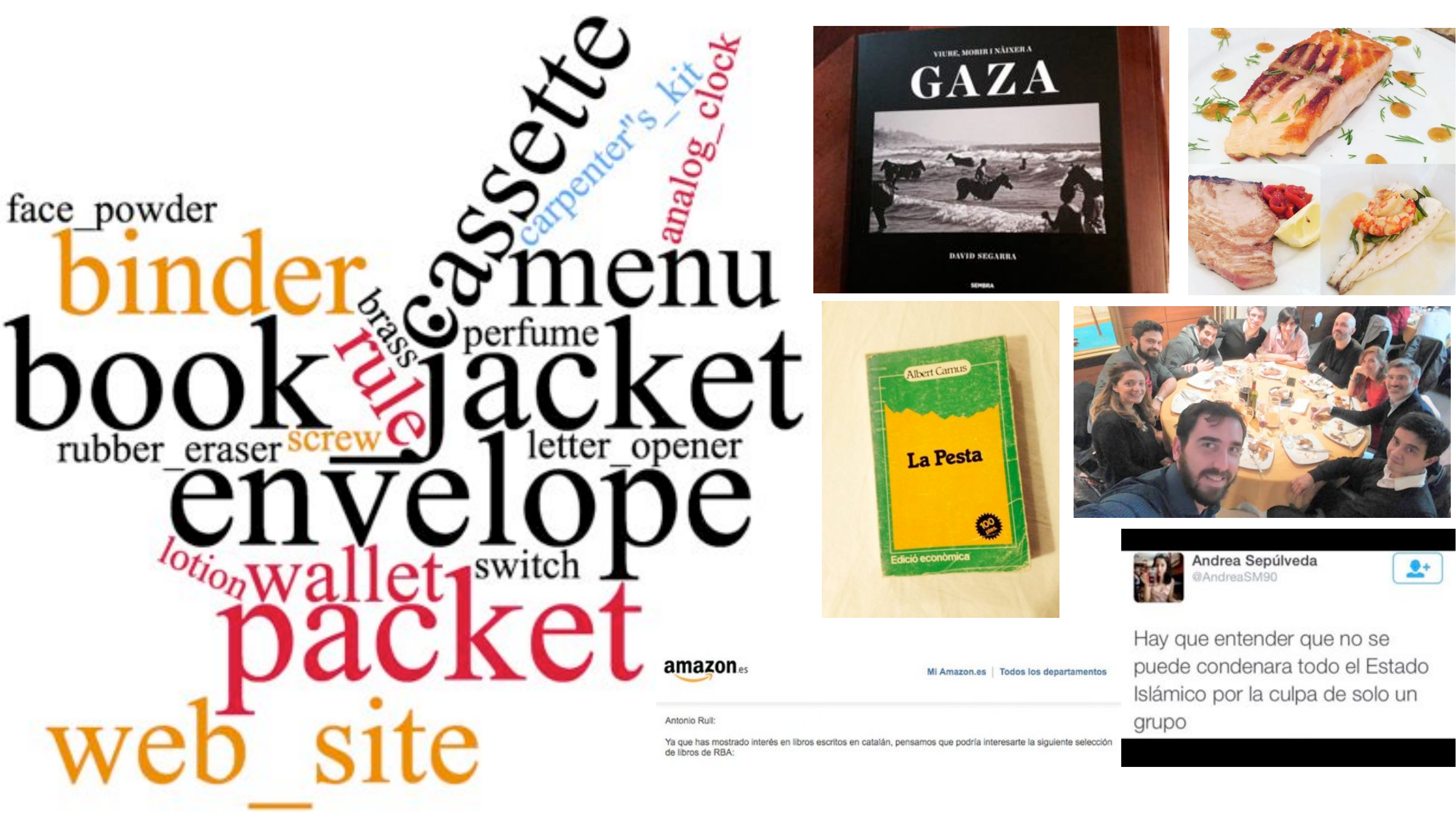}
        \caption{From 25-34 ages (Spanish).}
        \label{fig:es25}
    \end{subfigure}\\
    \begin{subfigure}[b]{0.4\textwidth}
        \includegraphics[width=\hsize]{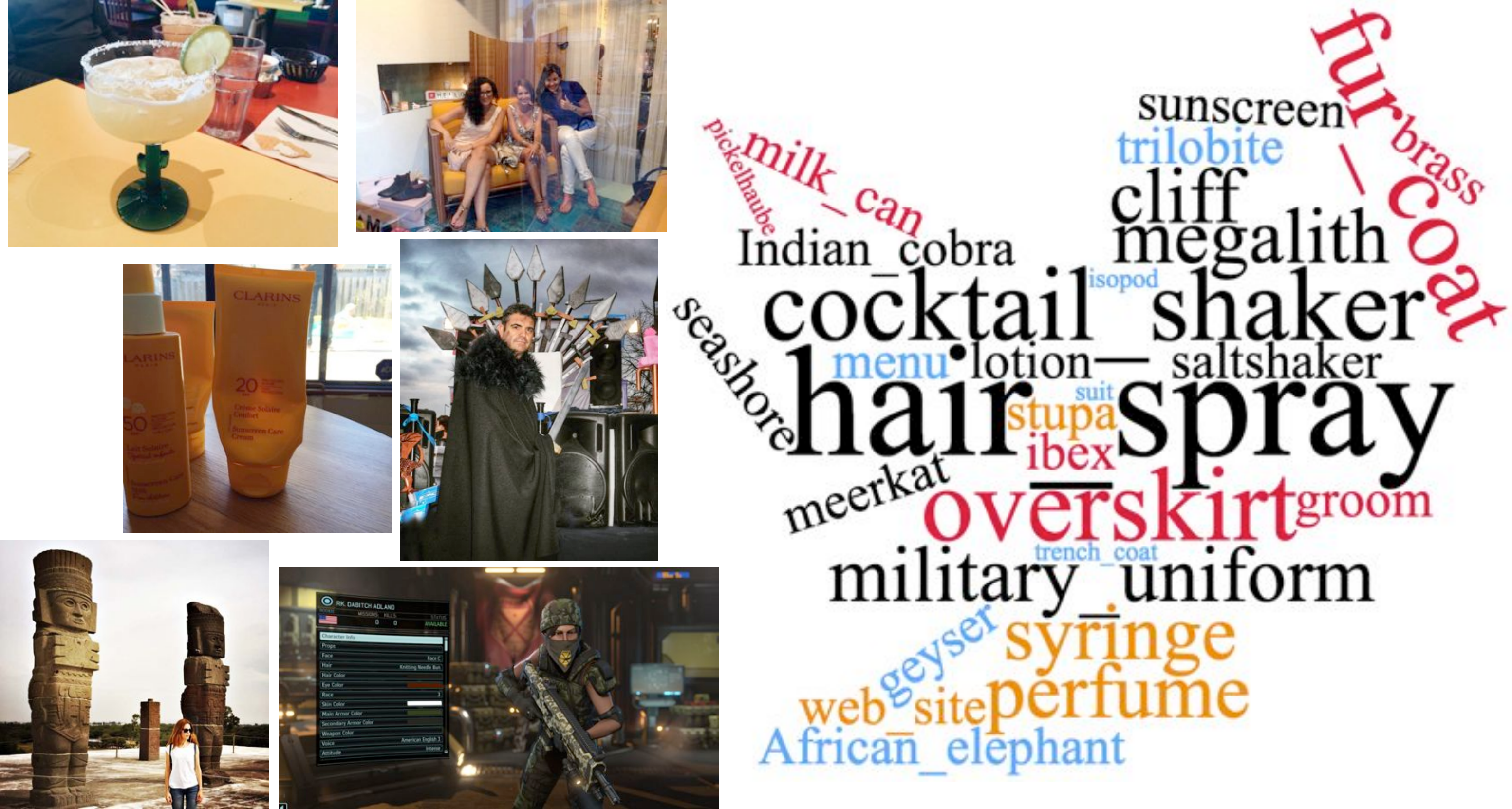}
        \caption{From 35-50 ages (English).}
        \label{fig:en35}
    \end{subfigure}
    \begin{subfigure}[b]{0.4\textwidth}
        \includegraphics[width=\hsize]{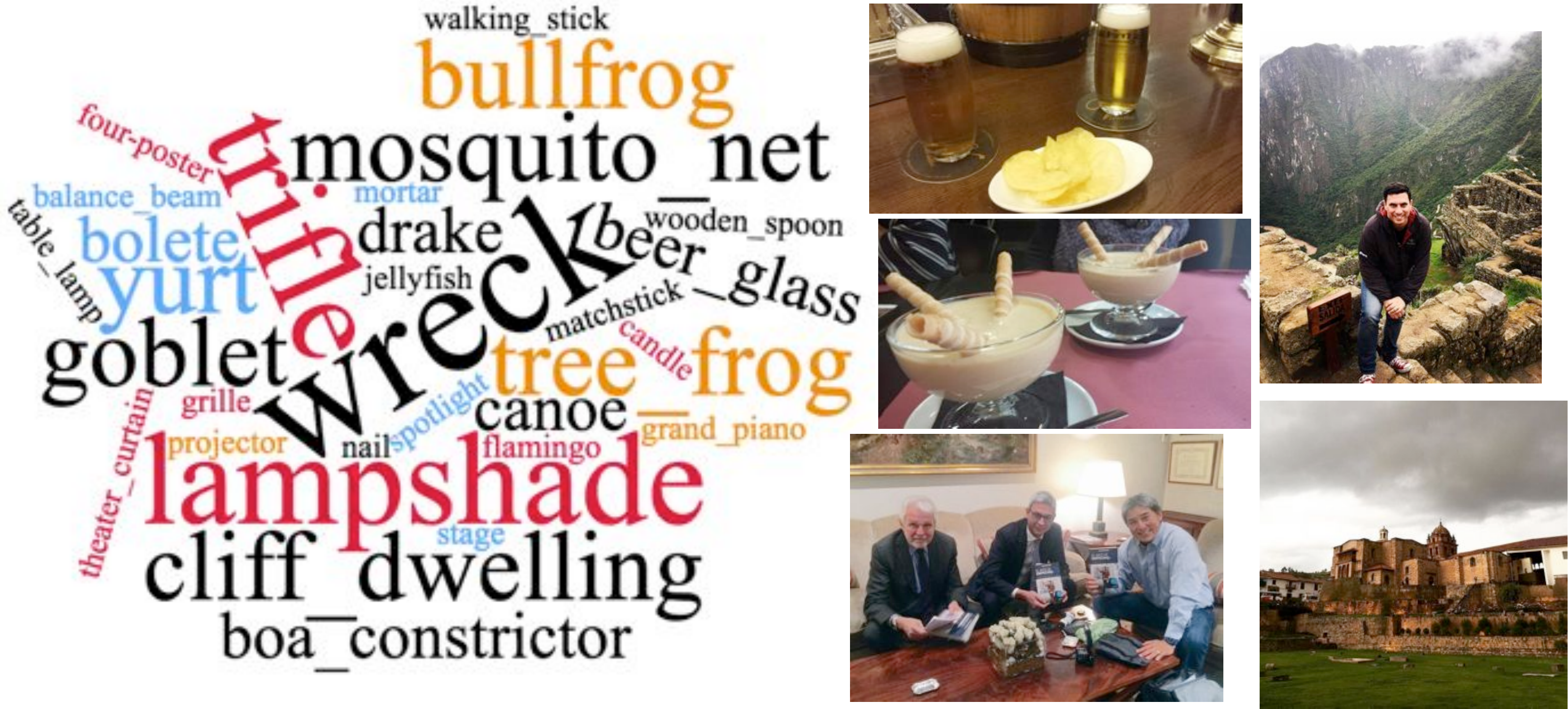}
        \caption{From 35-50 ages (Spanish).}
        \label{fig:es35}
    \end{subfigure}\\
    \begin{subfigure}[b]{0.4\textwidth}
        \includegraphics[width=\hsize]{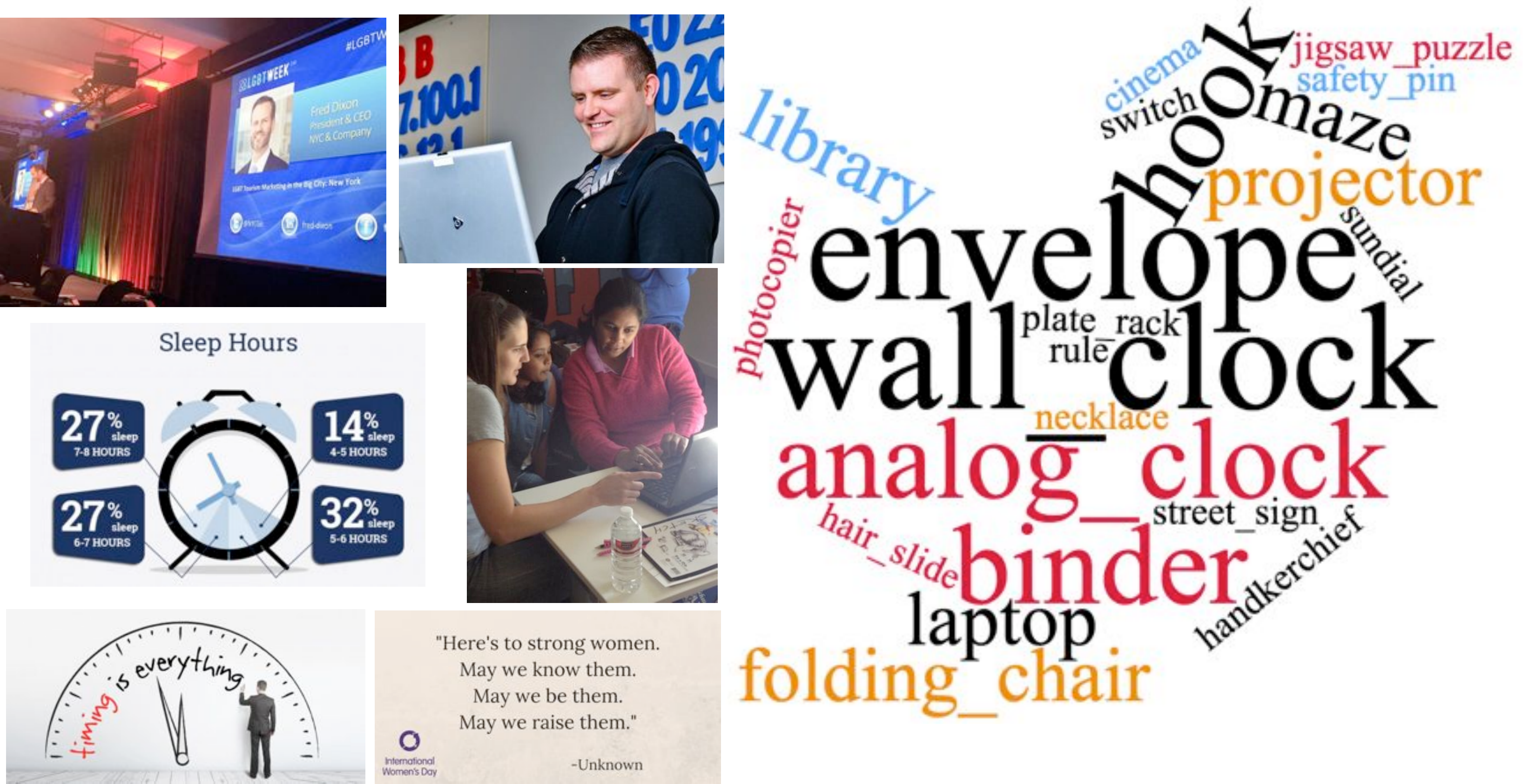}
        \caption{From 51-64 ages (English).}
        \label{fig:en50}
    \end{subfigure}
    \begin{subfigure}[b]{0.4\textwidth}
        \includegraphics[width=\hsize]{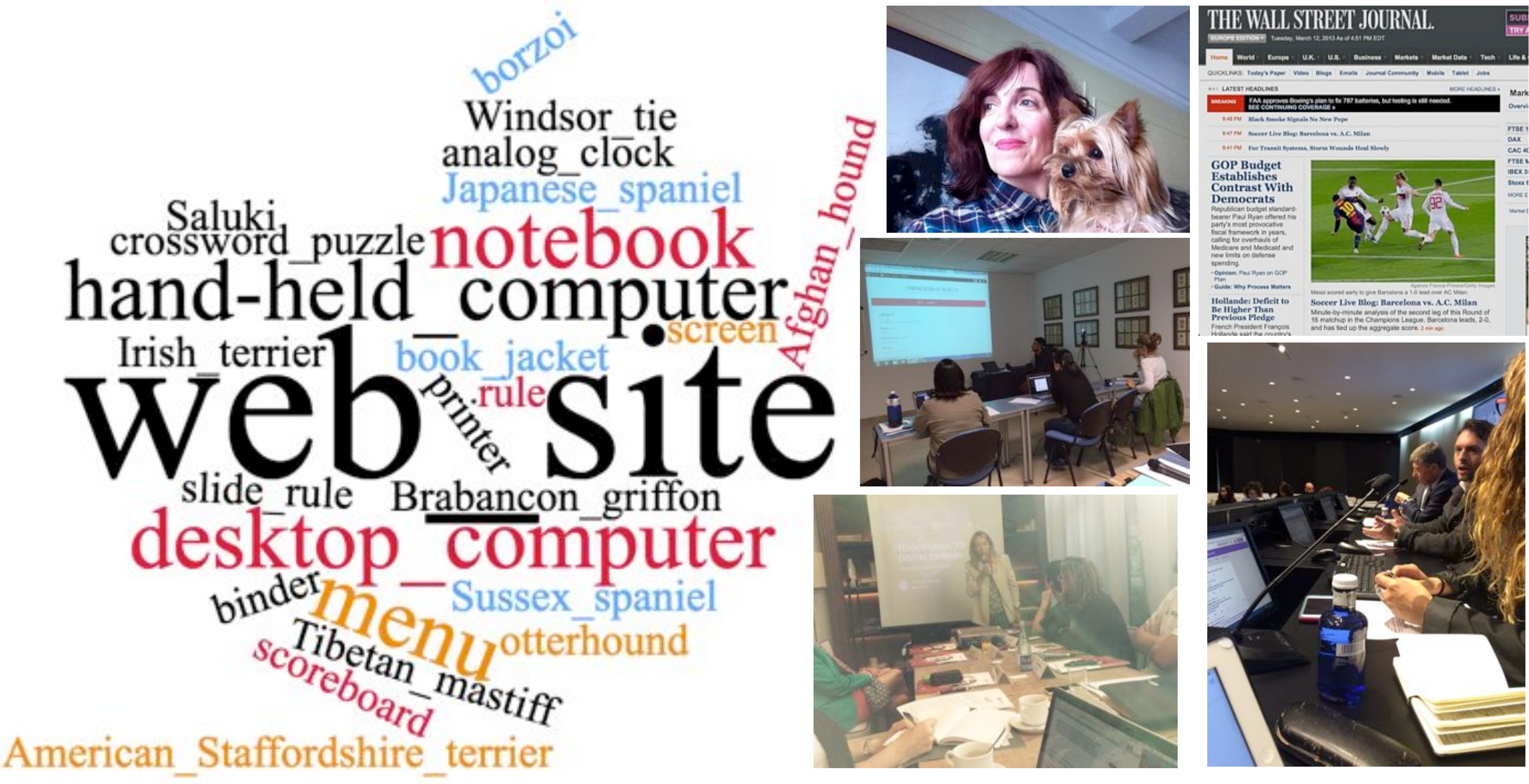}
        \caption{From 51-64 ages (Spanish).}
        \label{fig:es50}
    \end{subfigure}\\
    \begin{subfigure}[b]{0.4\textwidth}
        \includegraphics[width=\hsize]{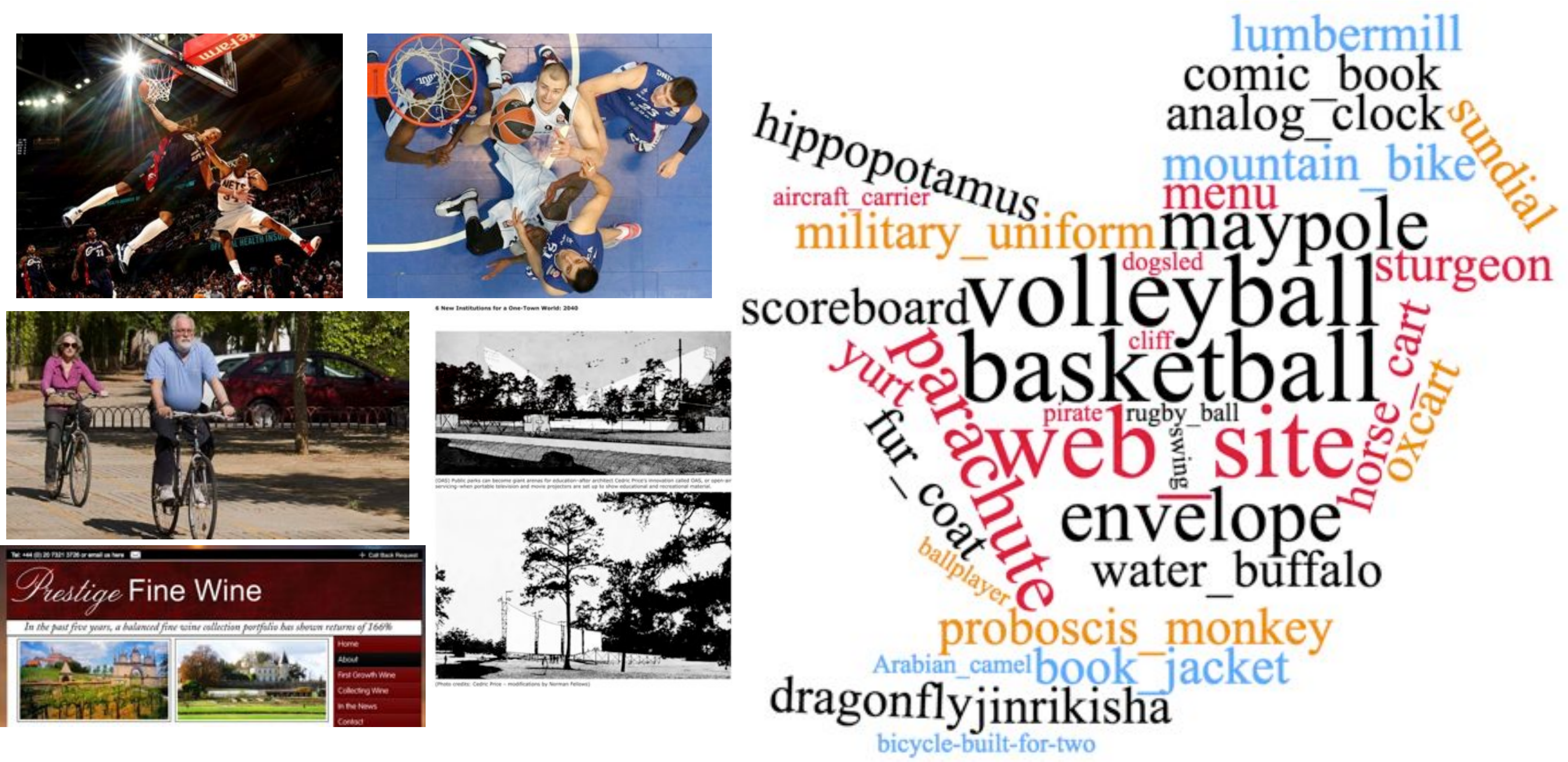}
        \caption{From 65-N ages (English).}
        \label{fig:en65}
    \end{subfigure}
    \begin{subfigure}[b]{0.4\textwidth}
        \includegraphics[width=\hsize]{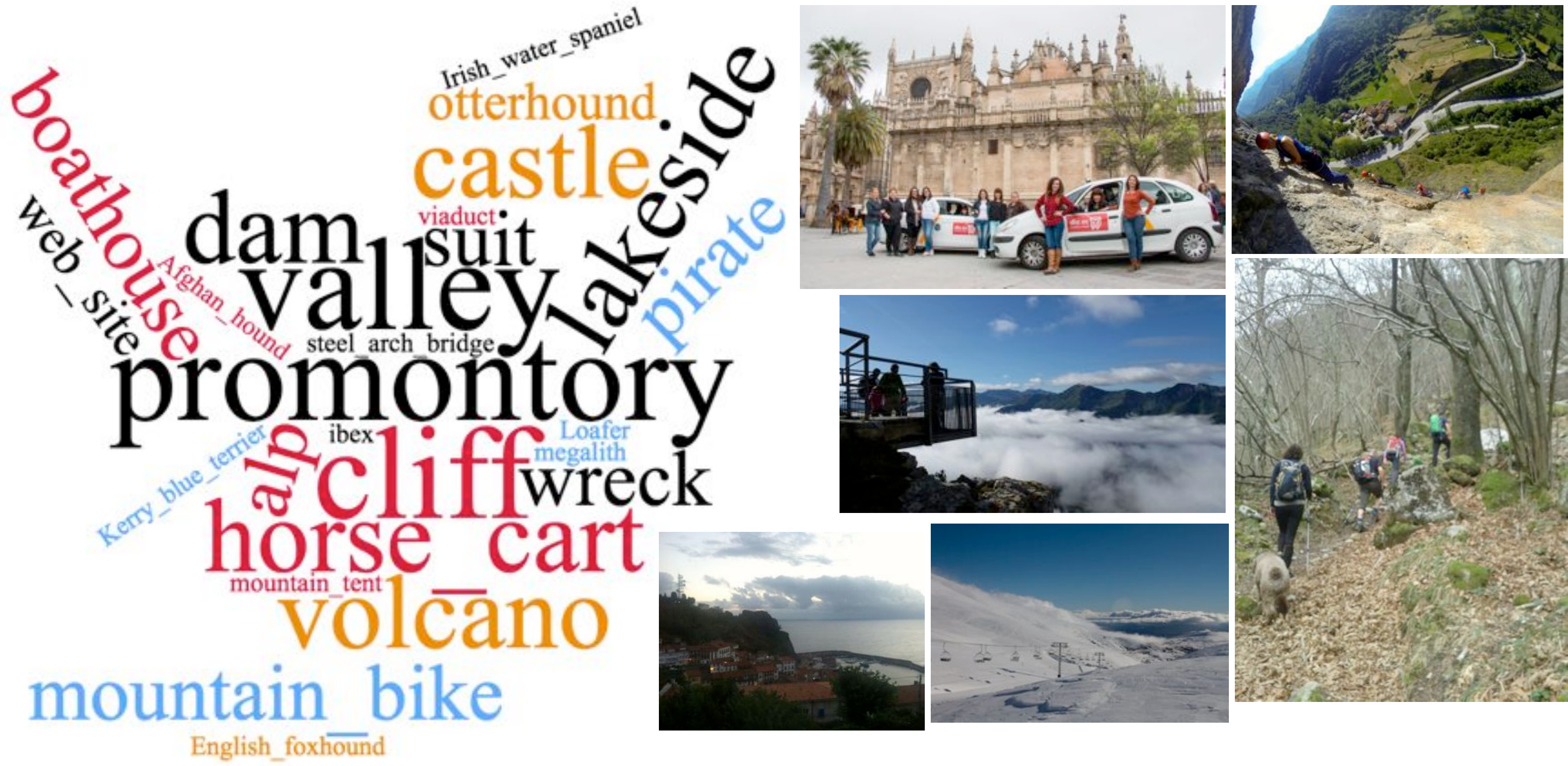}
        \caption{From 65-N ages (Spanish).}
        \label{fig:es65}
    \end{subfigure}
    \caption{Most frequent words, in addition to images posted among age ranges. On the right, from English corpus: (a), (c), (e), (g) and (i). On the left, from Spanish corpus: (b), (d), (f), (h) and (j). Size of the words indicates their frequency, being the smaller sizes the less frequent.}
    \label{words_most_labeled}
\end{figure*}

A second qualitative evaluation is presented in Figure~\ref{words_most_labeled}, this time showing the most posted words by age ranges in both corpora. For this purpose, each row presents an age range by a sample of images and its respectively cloud words. Here the size of the word indicates its frequency. The idea is to allow the reader to judge the visual information that we can extract from images posted in Twitter.

\section{Conclusions and Future Work}\label{conclus}

This paper explored the use of visual information to perform both age and gender identification in social media, specifically in Twitter. Novel methods for AP using visual information were proposed, as well as techniques based on multimodal techniques (text+images) for approaching the task. The models incorporating visual information rely on a CNN for feature extraction. The usefulness of images for AP was also explored by contrasting performance when using tweeted and retweeted images for the predictive models. 
Furthermore, 
we extended a benchmark data set for AP (PAN-2014 Twitter dataset) to include visual information by incorporating images from the users' profiles. The release of the later dataset is  an important contribution of this work to the state-of-the-art on AP, as it will motivate further research on visual and multimodal approaches to AP. 

Using the extended benchmark, we conducted an extensive evaluation including textual, visual and multimodal methods for  AP. The obtained results represent relevant evidence on the usefulness of visual information for AP. On the one hand, experimental results suggested that approaches based on multimodal information result in better performance, when compared to single modality approaches in both, age and gender prediction. On the other hand, results  indicated that images tend to be more relevant than text for determining the gender of Twitter users. We also found that the usefulness of visual information is somewhat dependent on the language of tweets. 

Regarding the analysis on the discriminative capabilities of tweeted and retweeted images, the obtained results did not allow us to formulate a convincing conclusion. However, results seemed to indicate that the image source matters. For example, results  showed that females gender identification was more accurate when using tweeted images, whereas for male gender identification using retweeted images worked better. 

For future work, we plan to extend this study by incorporating more advanced techniques for modeling both visual and textual information. We also consider evaluating the usefulness of other information modalities for AP, such as the posting behavior and video sharing.

\section*{Acknowledgments}
This research was supported by CONACyT under scholarships 401887, 214764 and 243957; project grants 247870, 241306 and 258588; and the Thematic Networks Program (Language Technologies Thematic Network, project 281795).

\bibliographystyle{plain}

\end{document}